\documentclass[letterpaper]{article} % DO NOT CHANGE THIS
\usepackage[submission]{aaai23}  % DO NOT CHANGE THIS
\usepackage{times}  % DO NOT CHANGE THIS
\usepackage{helvet}  % DO NOT CHANGE THIS
\usepackage{courier}  % DO NOT CHANGE THIS
\usepackage[hyphens]{url}  % DO NOT CHANGE THIS
\usepackage{graphicx} % DO NOT CHANGE THIS
\usepackage{natbib}  % DO NOT CHANGE THIS AND DO NOT ADD ANY OPTIONS TO IT
\usepackage{caption} % DO NOT CHANGE THIS AND DO NOT ADD ANY OPTIONS TO IT
\usepackage{algorithm}
\usepackage{algorithmic}

\usepackage{subcaption}
\usepackage{multirow}
\usepackage{pifont}
\usepackage{color}
\usepackage{amsmath,amsfonts,amssymb}

\usepackage{newfloat}
\usepackage{listings}
\DeclareCaptionStyle{ruled}{labelfont=normalfont,labelsep=colon,strut=off} % DO NOT CHANGE THIS
\floatstyle{ruled}
\newfloat{listing}{tb}{lst}{}
\floatname{listing}{Listing}
\title{X-ReID: Cross-Instance Transformer for Identity-Level Person Re-Identification}
\author {
    Leqi Shen\thanks{equal contribution}\textsuperscript{\rm 1,2},
    Tao He\footnotemark[1]\thanks{corresponding author}\textsuperscript{\rm 1,2},
    Yuchen Guo\textsuperscript{\rm 2},
    Guiguang Ding\footnotemark[2]\textsuperscript{\rm 1,2}
}
\affiliations {
    \textsuperscript{\rm 1} School of Software, Tsinghua University, Beijing, China \\
    \textsuperscript{\rm 2} Beijing National Research Center for Information Science and Technology (BNRist) \\
    \{lunarshen, kevin.92.he, yuchen.w.guo\}@gmail.com,
     dinggg@tsinghua.edu.cn
}
\begin{document}

\maketitle

\begin{abstract}
Currently, most existing person re-identification methods use \emph{Instance-Level} features, which are extracted only from a single image. However, these \emph{Instance-Level} features can easily ignore the discriminative information due to the appearance of each identity varies greatly in different images. Thus, it is necessary to exploit \emph{Identity-Level} features, which can be shared across different images of each identity. In this paper, we propose to promote \emph{Instance-Level} features to \emph{Identity-Level} features by employing cross-attention to incorporate information from one image to another of the same identity, thus more unified and discriminative pedestrian information can be obtained. We propose a novel training framework named X-ReID. Specifically, a Cross Intra-Identity Instances module (IntraX) fuses different intra-identity instances to transfer \emph{Identity-Level} knowledge and make \emph{Instance-Level} features more compact. A Cross Inter-Identity Instances module (InterX) involves hard positive and hard negative instances to improve the attention response to the same identity instead of different identity, which minimizes intra-identity variation and maximizes inter-identity variation. Extensive experiments on benchmark datasets show the superiority of our method over existing works. Particularly, on the challenging MSMT17, our proposed method gains 1.1$\%$ mAP improvements when compared to the second place.
% Specifically, we propose a novel training framework named X-ReID, which contains an Identity-to-Instance module to improve Instance-Level features and an Instance-to-Identity module to boost the Identity-Level ability. Both modules can further promote each other. In addition, we also design a post-processing method termed as AttnRank for inference to fuse multiple Instance-Level features to generate one unified and robust Identity-Level features, which can be adopted to optimize general Transformer-based ranking results. Extensive experiments on benchmark datasets show the superiority of our methods over existing works. Particularly, without bells and whistles, our proposed methods achieve 80.9$\%$ mAP on the challenging MSMT17, which surpass state-of-the-art performance by a large margin.
\end{abstract}

\section{Introduction}

\label{sec:intro}

Person Re-Identification (ReID) aims to retrieve a person of interest across multiple non-overlapping cameras. Given a single image of an identity, the goal of the ReID model is to learn the pedestrian characteristics, so that the distance of intra-identity features is smaller than that of inter-identity. However, the task of ReID is very challenging, since various complex conditions, such as human pose variations, different backgrounds, or varying illuminations, usually appear on different images of the same identity.

\begin{figure}[t]
\centering
\scalebox{0.7}
{
\includegraphics[width=\linewidth]{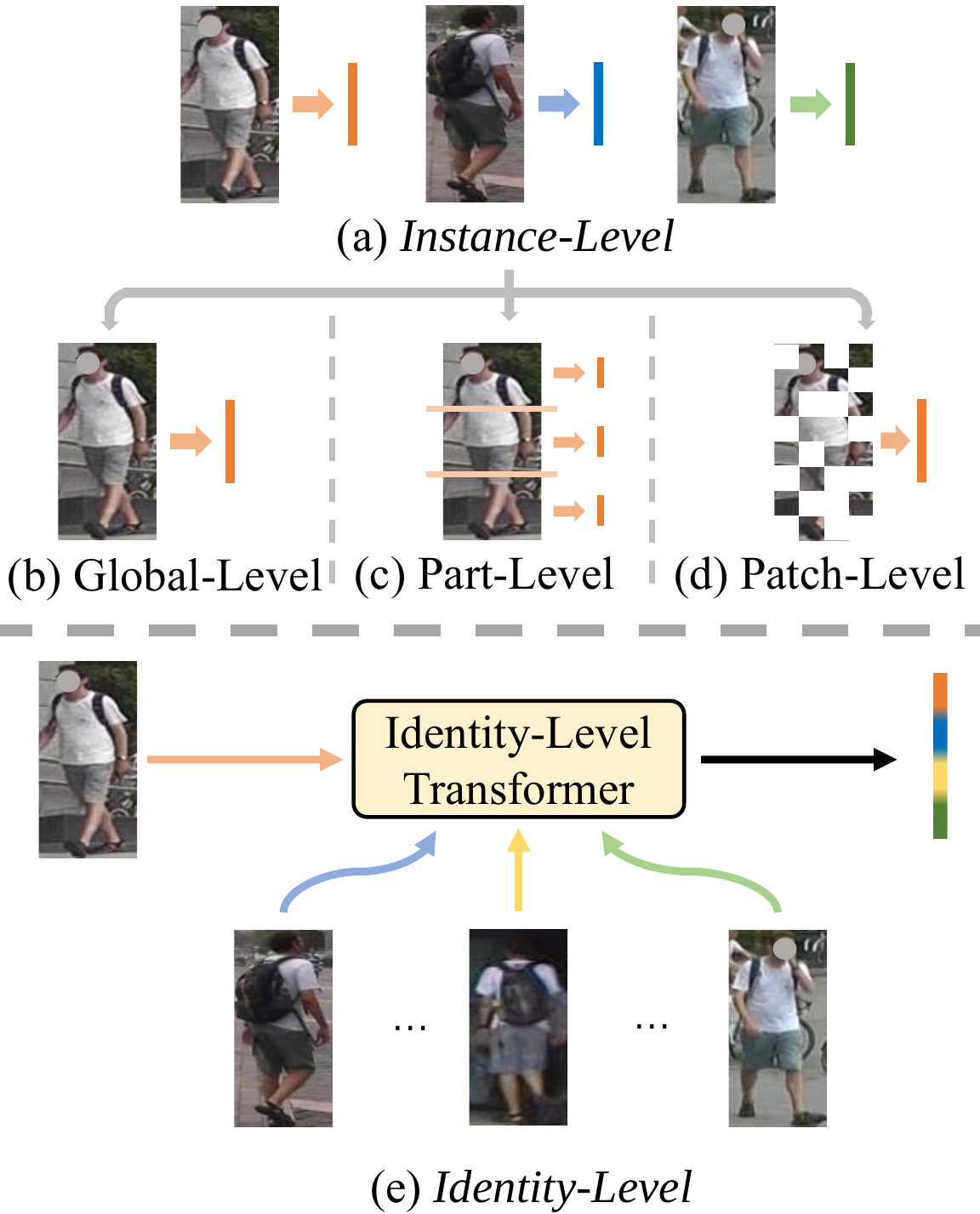}
}
\caption{The images on different sides are the same identity. (a) \emph{Instance-Level} methods can be classified into: (b) Global-Level methods, (c) Part-Level methods and (d) Patch-Level methods. As illustrated in (e), our proposed method exploits \emph{Identity-Level} features shared across different images of the same identity.}
\label{fig:intro}
\end{figure}
% The images from different sides are the same identity. 
% our proposed method exploits Identity-Level features shared across different images of the same identity.

Several previous studies\cite{ye2021deep,luo2019bag,chen2019abd,sun2018beyond,zhu2020identity,he2021transreid} present promising results on supervised ReID. Since these works extract features from only a single image, rather than from a combination of multiple images of the same identity, we refer to them as \emph{Instance-Level} methods. According to the different processing methods for individual images, Instance-Level methods can be roughly classified into the following categories. 
(1) Global-Level methods\cite{luo2019bag,chen2020salience,chen2019abd,zhang2020relation,fang2019bilinear,si2018dual,zheng2019re,chen2018group,luo2019spectral}, which directly learn the global representation by optimizing Instance-Level features of each single complete image, as shown in Figure~\ref{fig:intro}.b.
(2) Part-Level methods\cite{sun2018beyond,wang2018learning,li2021diverse,zhu2020identity,he2020guided,zhang2019densely,jin2020semantics,zhu2021aaformer, zhu2022dual}, as shown in Figure~\ref{fig:intro}.c, which learn local aggregated features from different parts.
(3) Patch-Level methods such as TransReID\cite{he2021transreid}, which rearranges and re-groups the patch embeddings of each single image (see Figure~\ref{fig:intro}.d). Each group forms an incomplete global representation. 
% without interaction between multiple images of the same identity

Due to various factors such as human pose and camera variations, the appearance of the same identity varies greatly in different images. Instance-Level methods focus on features only from a single image, so the features of different images of the same identity are prone to large differences. For example, Figure~\ref{fig:intro}.a shows the front and back of the same identity captured by different cameras. As the backpack in the back image is more salient than that in the front image, features independently learned by Instance-Level methods will vary greatly. Consequently, it may lead to intra-identity variation being larger than inter-identity variation, which further limits the performance in inference.

% We argue the reason is that Instance-Level methods are unable to learn discriminative features shared across different images of the same identity. To resolve this problem, we propose a novel Identity-Level method, as shown in  Figure~\ref{fig:intro}.e. For the discriminative information of the ``backpack'' in the back image of $A$, the front of $A$ can gain knowledge from the back to strengthen its representation, thereby reducing intra-identity variation. With the back of hard negative $B$, the difference between hard positive and negative will be learned, so that inter-identity variation is increased further. In this way, more unified and robust pedestrian information can be learned.

We argue the reason is that Instance-Level methods are unable to learn discriminative features shared across different images of the same identity. To resolve this problem, we propose a novel \emph{Identity-Level} method to fuse different Instance-Level features of the same identity, as shown in  Figure~\ref{fig:intro}.e. In this way, more unified and robust pedestrian information can be learned. With our method, for the discriminative information of the ``backpack'' in the back image, the front can gain knowledge from the back to strengthen its representation, thereby reducing intra-identity variation, enlarging inter-identity variation and obtaining a more robust features.

% \begin{figure}[t]
% \centering
% \includegraphics[width=\linewidth]{Fig/Intro_example.pdf}
% \caption{A toy example of Instance-Level Transformer and Identity-Level Transformer.}
% \label{fig:intro_example}
% \end{figure}

Our proposed method is based on the pure Transformer structure\cite{dosovitskiy2020image}. The Multi-Head Attention module can exploit long-range dependencies between patches divided from the input image, which can get fine-grained features with rich structural patterns. However, previous works\cite{he2021transreid,zhu2021aaformer} only use the Instance-Level Transformer and thus can not jointly encode multiple images to extract Identity-Level features. In this paper, we introduce cross-attention to the encoder, which has not been fully explored in vision task, especially in ReID. Cross-attention can incorporate information from different instances, which mitigates the impact of the variable appearance of the same identity. 

Based on the attention mechanism, we introduce a novel training framework, CrossReID termed as X-ReID, which performs attention across different instances of the same identity to extract Identity-Level features. Firstly, we propose a Cross Intra-Identity Instances module (IntraX). Cross-attention is applied among different intra-identity instances to get more unified features. Then, we transfer Identity-Level knowledge to Instance-Level features, which makes features more compact. Second, We further propose a Cross Inter-Identity Instances module (InterX). Hard positive and negative instances are utilized in cross-attention. The fused features, which contain hard negative instance, are less discriminative. The fused features are optimized by our proposed X-Triplet loss to improve the attention response to the same identity rather than different identity, so that intra-identity variation is minimized and inter-identity variation is maximized. Our modules consider both Intra- and Inter-Identity instances to improve the Identity-Level ability of Instance-Level features. Only the improved Instance-Level features are used for inference, which brings no additional cost compared with Transformer-based state-of-the-art methods.

% instead of all positive instances in IntraX
% Firstly, we propose an i\textbf{D}entity-\textbf{to}-in\textbf{S}tance module (D2S). The D2S builds an exponential moving average layer to transfer Identity-Level knowledge to Instance-Level features. Next, we propose an in\textbf{S}tance-\textbf{to}-i\textbf{D}entity module (S2D). Different Instance-Level features are fused at a Multi-Head Cross-Attention sub-layer to boost the Identity-Level ability. Both modules can complement each other to promote Instance-Level features to Identity-Level features, which have more unified and discriminative pedestrian information. By default, only Instance-Level features are used for inference, which brings no additional cost compared with Transformer-based state-of-the-art methods.

% In addition to our proposed training framework, we also design a post-processing method for inference, Attention Rank termed as AttnRank, to optimize general Transformer-based ReID ranking results. First of all, we adopt the $k$-fixed-reciprocal nearest neighbors to get a fixed-length neighbor set. Then, we fuse these features at the last Transformer layer to yield new Identity-Level features, which are used as the final features for pedestrian retrieval. Besides, our proposed AttnRank can be combined with the widely-used post-processing method ReRank\cite{zhong2017re}, to further improve the ReID performance.

Our major contributions can be summarized as follows:
\begin{itemize}
\item We propose to exploit Identity-Level features shared across different images of each identity to build a more unified, discriminative and robust representation.
\item We introduce a novel training framework, termed as X-ReID, which contains IntraX for Intra-Identity instances and InterX for Inter-Identity instances to improve the Identity-Level ability of Instance-Level features.
\item The overall method is evaluated on benchmark datasets, MSMT17\cite{wei2018person} and Market1501\cite{zheng2015scalable}. Extensive experimental results validate the effectiveness of Identity-Level and X-ReID achieves state-of-the-art performance.
\end{itemize}

\section{Related Work}

\subsection{Supervised ReID}

Existing works extract instance representations from a single image, so we refer to them as Instance-Level methods, which can be roughly categorized into three classes: (1) Global-Level methods\cite{zheng2017person,luo2019bag}, which learn the global representation of each single complete image. The misalignment problem is handled in attention-based methods\cite{chen2020salience,chen2019abd,zhang2020relation,fang2019bilinear,si2018dual,zheng2019re,chen2018group,luo2019spectral}. (2) Part-Level methods\cite{sun2018beyond,wang2018learning,li2021diverse,zhu2020identity,he2020guided,zhang2019densely,jin2020semantics,zhu2021aaformer}, which learn local aggregated features from different parts. (3) Patch-Level methods such as TransReID\cite{he2021transreid}, which shuffles patches into several incomplete representations.

There are a few CNN-based works that leverage the cross-image information. These works mine the relationships between multiple Instance-Level feature maps. DuATM\cite{si2018dual} performs dually attentive comparison for pair-wise feature alignment and refinement. CASN\cite{zheng2019re} designs a Siamese network with attention consistency. Group similarity is proposed in GCSL\cite{chen2018group} for learning robust multi-scale local similarities. SFT\cite{luo2019spectral} optimizes group-wise similarities to capture potential relational structure.

\begin{figure*}[t]
\centering
\scalebox{0.8}
{
\includegraphics[width=\linewidth]{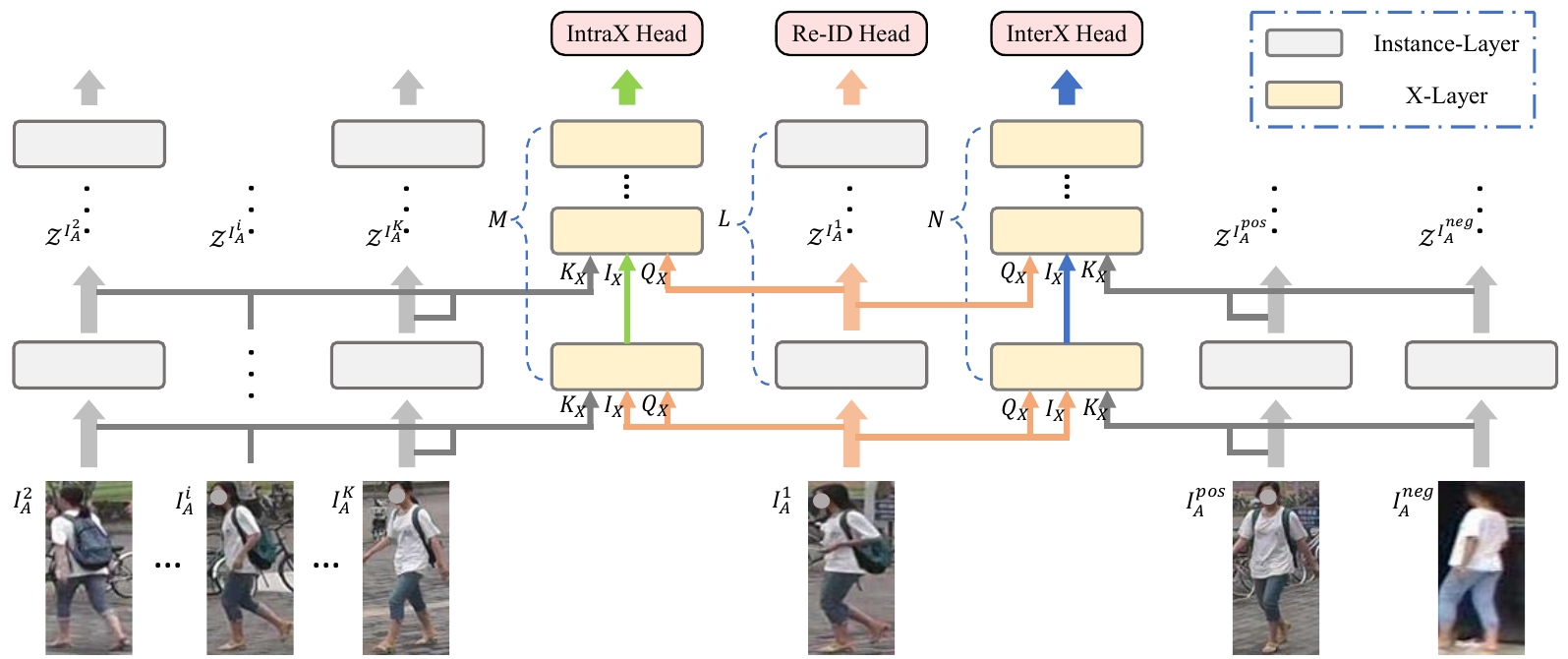}
}
\caption{Framework of our proposed X-ReID, based on $L$ layers Instance-Level Transformer. It is composed of a Cross Intra-Identity Instances module (IntraX) and a Cross Inter-Identity Instances (InterX). We build $M$ X-Layers for IntraX, $N$ X-Layers for InterX, and set $M=N=L$. The details of X-Layer are shown in Figure~\ref{fig:xlayer}.}
\label{fig:X-ReID}
\end{figure*}

\subsection{Transformer in Vision}

Inspired by the success of Transformers\cite{vaswani2017attention} in natural language tasks, many studies\cite{khan2021transformers,han2020survey,gao2021ts,carion2020end,xie2021segformer,zhao2021point,chen2021pre} have shown the superiority of Transformer in vision tasks. ViT\cite{dosovitskiy2020image} first introduces a pure Transformer to image classification. 

In the field of ReID, PAT\cite{li2021diverse} utilizes Transformer to improve CNN backbone\cite{he2016deep}. NFormer\cite{wang2022nformer} proposes a parametric post-processing method by modeling the relations among all the input images. TransReID\cite{he2021transreid} investigates a pure Transformer encoder framework, and AAformer\cite{zhu2021aaformer} integrates the part alignment into the self-attention.  DCAL\cite{zhu2022dual} introduces cross-attention mechanisms to learn subtle feature embeddings. Instead, we exploit cross-attention to fuse Instance-Level features. 

As for image classification, cross-attention in Cross-ViT\cite{chen2021crossvit} is used to learn multi-scale features. In Trear\cite{li2021trear}, features from different modalities are fused to obtain the conjoint cross-modal representation. Unlike these approaches, our proposed cross-attention extracts Identity-Level features for more unified and discriminative pedestrian information.

\section{Method}

Our method is built on top of pure Transformer architecture, so we first present a brief overview of Instance-Level Transformer in Sec~\ref{sec:instance}. Then, we introduce the proposed Identity-Level framework (X-ReID) in Sec~\ref{sec:xreid}, which contains a Cross Intra-Identity Instances module (IntraX) and a Cross Inter-Identity Instances (InterX).

\subsection{Instance-Level Transformer for ReID}
\label{sec:instance}

\begin{figure}
\centering
\includegraphics[width=\linewidth]{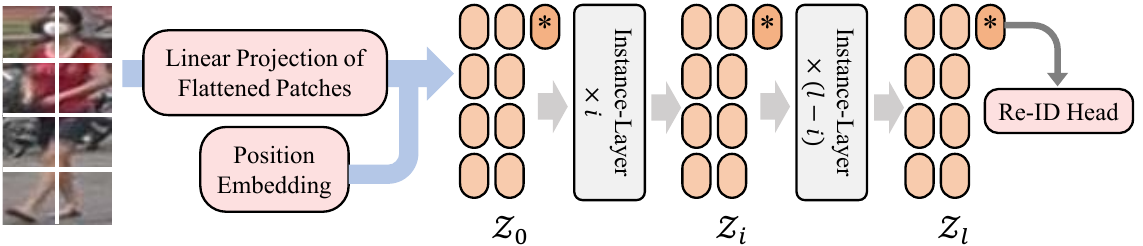}
\caption{Framework of Instance-Level Transformer. $\ast$ refers to CLS token.}
\label{fig:instance}
\end{figure}

Instance-Level Transformer for ReID is a Transformer-based strong baseline and we follow the baseline in TransReID\cite{he2021transreid}. As shown in Figure~\ref{fig:instance}, given a single input image, we reshape it into a sequence of flattened patches $N \times (P^{2} \times C)$, where C is the number of channels. $P$ is the resolution of each patch, and $N$ is the length of fixed-sized patches according to image size. We flatten the patches and map to vectors with a trainable linear projection. An extra learnable embedding CLS is added to the sequence of embedded patches. Learnable position embeddings are used for retaining spatial information. The output embeddings can be expressed as: $\mathcal{Z}_{0}=[{z}_{cls}^{0};{z}_{1}^{0};...;{z}_{N}^{0}]$, where $\mathcal{Z}_{0}$ represents the input fed into backbone, ${z}_{cls}$ is CLS token and ${z}_{patch}=\left \{ {z}_{i} | i=1,...,N \right \}$ are patch tokens. 

The backbone is composed of a stack of $L$ identical Transformer layers. Because the input is a single image and the layers encode the Instance-Level information, the Transformer layer is named as Instance-Layer in this paper. And each layer can encode the embeddings: $\mathcal{Z}_{i}=f(\mathcal{Z}_{i-1} | {\theta}_{Ins}^{i})$, where $\mathcal{Z}_{i}$ is the output of the $i$-th Instance-Layer and ${\theta}_{Ins}^{i}$ is the weight of this layer. After LayerNorm\cite{ba2016layer}, the last layer’s output of CLS token ${z}_{cls}^{l}$ gets Instance-Level features ${f}_{Ins}$, to extract global representation, which is used to calculate ID loss\cite{zheng2017discriminatively} $\mathcal{L}_{ID}$ and Triplet loss\cite{liu2017end} $\mathcal{L}_{Tri}$ that constitute the ReID Head\cite{luo2019bag}:

\begin{equation}
    \mathcal{L}_{Ins} = \mathcal{L}_{ID}(f_{Ins}) +  \mathcal{L}_{Tri}(f_{Ins})
\label{formula:ReID_Head}
\end{equation}

\subsection{X-ReID}
\label{sec:xreid}

Our proposed Identity-Level framework is illustrated in Figure~\ref{fig:X-ReID} based on Instance-Level Transformer, which contains IntraX and InterX. For a single image $I_A^1$ of person $A$, the tokens obtained from Instance-Level Transformer are $\mathcal{Z}^{I_A^1}=\left \{ \mathcal{Z}^{I_A^1}_{i} | i=0,1,...,L \right \}$. On the left side of Figure~\ref{fig:X-ReID}, as for other images of the same person $A$, we get $\left \{\mathcal{Z}^{I_A^i} | i=2,...,K\right \}$. IntraX are performed among Intra-Identity instances. $\left \{\mathcal{Z}^{I_A^i}\right \}$ as Key and $\mathcal{Z}^{I_A^1}$ as Query are fed to X-Layers to extract Identity-Level features $f_{IntraX}$ of person $A$. Rather than all positive instances as IntraX does, InterX utilizes both positive instance $I_A^{pos}$ and negative instance $I_A^{neg}$. As shown on the right side of Figure~\ref{fig:X-ReID}, $\mathcal{Z}^{I_A^{pos}}$ and $\mathcal{Z}^{I_A^{neg}}$ are used as Key and we get features $f_{InterX}$. Our X-ReID explores both Intra- and Inter-Identity instances to enhance Instance-Level features.

\begin{figure}
\centering
\scalebox{0.80}
{
\includegraphics[width=\linewidth]{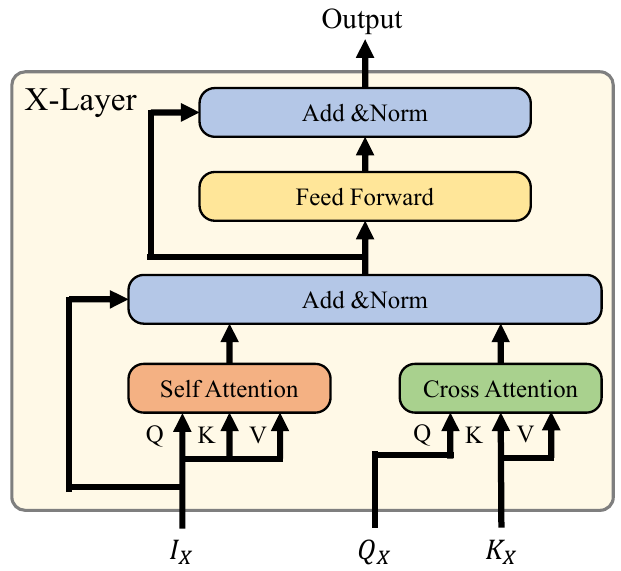}
}
\caption{Illustration of X-Layer. In both IntraX and InterX, $Q_X$ are the same. In IntraX, $K_X$ are all positive instances in the mini-batch. As for InterX, $K_X$ are hard positive and hard negative. $I_X$ are usually the output of the previous Layer.}
\label{fig:xlayer}
\end{figure}

\subsubsection{X-Layer.}

Instance-Layer, which consists of alternating layers of \textbf{M}ulti-\textbf{H}ead \textbf{A}ttention and FFN blocks, only has Self Attention. The output of MHA is formulated as:

\begin{equation}
    {\rm MHA}(Q,K,V) = {\rm softmax}(\frac{QK^{T}}{\sqrt{d_k}})V
\label{formula:MHA}
\end{equation}

\noindent where ${d}_{k}$ is the scaling factor. Figure~\ref{fig:xlayer} shows the detailed structure of our proposed X-Layer. Cross Attention is added to Instance-Layer to become X-Layer. With Cross Attention, X-Layer can obtain features across multiple inputs:

\begin{equation}
\begin{split}
    {\rm Attention}_X(I_X,Q_X,K_X) & = {\rm MHA}(I_X, I_X, I_X) \\
    & + {\rm MHA}(Q_X, K_X, K_X)
\end{split}
\end{equation}

\noindent where ${\rm MHA}(I_X, I_X, I_X)$ is Self Attention and ${\rm MHA}(Q_X, K_X, K_X)$ is Cross Attention. It is worth noting that Self Attention and Cross Attention share parameters, so the parameters of X-Layer are the same as those of Instance-Layer. 
% After the $l$-th Instance-Layer, we get Instance-Level features ${f}_{Ins}$. Different image ${I}^{'}$ has the same identity as $I$, and $\mathcal{Z}_{l-1}^{'}$ is the output of the $(l\raisebox{0mm}{-}1)$-th Instance-Layer. $\mathcal{Z}_{l-1}$ and $\mathcal{Z}_{l-1}^{'}$ are fed into D2S-Layer and S2D-Layer, respectively. Therefore, Identity-Level outputs are D2S features ${f}_{D2S}$ and S2D features ${f}_{S2D}$, which are devised to improve our model.

% Due to the mini-batch size, it is infeasible to obtain Identity-Level features across all images of each identity. For each single image, we randomly sample an image with the same identity in a mini-batch. During our proposed training process, the fused features between random paired images will gradually become Identity-Level features.

\subsubsection{Cross Intra-Identity Instances.}
\label{sec:intraX}

We propose a Cross Intra-Identity Instances module to transfer Identity-Level knowledge to Instance-Level features. All positive instances of the input $I_A^1$ are fused in X-Layer to extract Identity-Level features. To enhance the Identity-Level capability, IntraX trains Instance-Level features to predict Identity-Level representation of the same identity by knowledge distillation.
% The design idea of D2S-layer is to obtain robust Identity-Level features by fusing original CLS token with patch tokens of another image on the existing backbone.

$M$ layers of X-Layer in IntraX uses the \textbf{E}xponential \textbf{M}oving \textbf{A}verage (EMA) on Instance-Layers, which is proposed to maintain consistency. Specifically, the parameters of $i$-th X-Layer can be calculated as: 

\begin{equation}
{\theta}_{IntraX}^{i} \gets \lambda \times {\theta}_{IntraX}^{i} + (1-\lambda) \times {\theta}_{Ins}^{i}
\end{equation}

\noindent where $\lambda$ follows a cosine schedule from 0.999 to 1 during training. 
 Knowledge about the different appearances of person $A$ is mixed in X-Layer. $I_X$, $Q_X$ and $K_X$ of $i$-th X-Layer in IntraX are:

\begin{equation}
\begin{split}
I_X^i & = {\rm Attention_X}(I_X^{i-1}, Q_X^{i-1}, K_X^{i-1}) \\
Q_X^i & = \mathcal{Z}_i^{I_A^1}, K_X^i = [\mathcal{Z}_i^{I_A^2}||...||\mathcal{Z}_i^{I_A^K}]
\end{split}
\label{formula:intrax}
\end{equation}

\noindent where $I_X^{i-1}$ are the outputs of the previous X-Layer, except that $I_X$ of the first layer are the same as $Q_x=\mathcal{Z}_0^{I_A^1}$. We concatenates all features of  positive instances of $I_A^1$ in the mini-batch to get $K_X \in \mathbb{R}^{(K-1)C}$. On account of X-Layer, IntraX incorporates different appearances of the same identity into Identity-Level features $f_{IntraX}$. Then, we transfer the knowledge from ${f}_{IntraX}$ to ${f}_{Ins}$:

\begin{equation}
\begin{split}
P(f)^{(i)} & = \frac{{\rm exp}(f^{(i)}/\tau)}{\sum\limits_{d=1}^{C} {\rm exp}(f^{(d)}/\tau)} \\{\mathcal{L}}_{IntraX} & = -{\sum\limits_{d=1}^{C} P({{f}_{IntraX}})^{(d)}{\rm log}P({{f}_{Ins}})^{(d)}}
\end{split}
\end{equation}

\noindent where $C=768$ are the dimensions of features and $\tau=0.05$ is a temperature parameter. The output of a softmax of the dot products of $f$ is the probability distribution $P$, which provides information on how the features represent knowledge. Matching distributions of $f_{Ins}$ and $f_{IntraX}$ by minimizing $\mathcal{L}_{IntraX}$ is used to improve Instance-Level features by Identity-Level features. Therefore, Instance-Level features are more compact.

\subsubsection{Cross Inter-Identity Instances.}
\label{sec:interX}

Our proposed Cross Inter-Identity Instances module deals with inter-identity instances of $I_A^1$, instead of all positive instances in IntraX. InterX uses both hard positive and hard negative as $K_X$ of X-Layer, where the negative instance $I_A^{neg}$ has not the same identity as the anchor $I_A^1$. X-Layers in InterX may incorrectly notice the negative, whose interfering information affects the final fused features of InterX. We optimize the fused features to suppress the attention response of the anchor to negative instance and improve the attention response to positive instance, so that intra-identity variation is minimized and inter-identity variation is maximized. 

Particularly, InterX contains $N$ layers of X-Layer that are learnable. $I_X$ and $Q_X$ are the same as in Equation~\ref{formula:intrax}. $K_X$ of $i$-th X-Layer in interX is:

\begin{equation}
K_X^i = [\mathcal{Z}_i^{I_A^{pos}}||\mathcal{Z}_i^{I_A^{neg}}]
\label{formula:interx}
\end{equation}

\noindent where $K_X^i$ is the result of concatenating $\mathcal{Z}_i^{I_A^{pos}}$ of hard positive $I_A^{pos}$ and $\mathcal{Z}_i^{I_A^{neg}}$ of hard negative $I_A^{neg}$. We conduct hard-mining\cite{hermans2017defense} in $\mathcal{L}_{Tri}$ to get hard positive and negative instance. As illustrated in Figure~\ref{fig:xtriplet}.a,  $\mathcal{L}_{Tri}$ in the ReID Head is calculated as follows:

\begin{equation}
\mathcal{L}_{Tri}(a,p,n) = {\rm log} [1+{\rm exp}(|| f_{a}-f_{p}) ||_2^2-|| f_{a}-f_{n}) ||_2^2]
\label{formula:triplet}
\end{equation}

\noindent where a, p and n is the anchor, hard positive and hard negative. The anchor may be closer to hard negative compared to hard positive. Therefore in Figure~\ref{fig:xtriplet}.b, the anchor as $Q_X$ is difficult to distinguish which one is positive in InterX. The fused features will be heavily influenced by hard negative. Red represents hard negative and green represents hard positive. In the fused features $f_{InterX}$, red occupies most of the space. Then, we propose X-Triplet loss:

\begin{figure}
\centering
\scalebox{0.75}
{
\includegraphics[width=\linewidth]{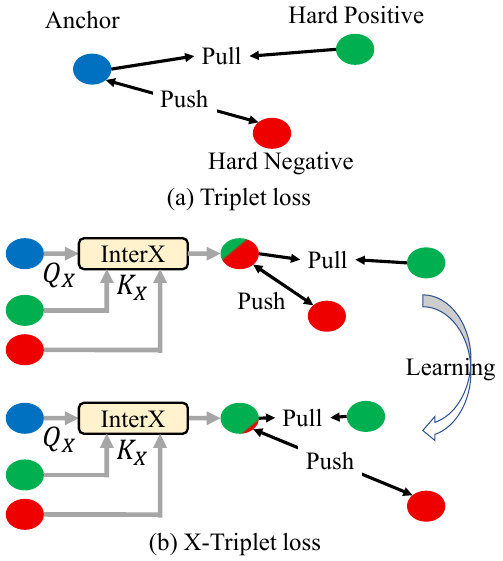}
}
\caption{(a) Hard-mining in Triplet loss selects the hardest positive and the hardest negative instances within the batch. (b) X-Triplet loss optimize the fused features from hard positive and negative. Hard positive will have more attention responses than hard negative in X-Layer.}
\label{fig:xtriplet}
\end{figure}

\begin{equation}
\mathcal{L}_{XTri}(f_{InterX}) = {L}_{Tri}(f_{InterX},p,n)
\label{formula:Xtriplet}
\end{equation}

\noindent Therefore, $f_{InterX}$ is pulled into hard positive and pushed away by hard negative. Then, the InterX Head composed of $\mathcal{L}_{ID}$ and $\mathcal{L}_{XTri}$ are shown as follows:

\begin{equation}
\mathcal{L}_{InterX} = {L}_{ID}(f_{InterX}) + {L}_{XTri}(f_{InterX})
\label{formula:InterXhead}
\end{equation}

\noindent By $\mathcal{L}_{InterX}$, the anchor are encouraged to pay more attention to hard positive than hard negative in X-Layer. $f_{InterX}$ will be optimized to represent the person of the anchor and hard positive. At this stage, green occupies most of the space. Accordingly, InterX simultaneously reduces the distance of intra-identity and enlarges the distance of inter-identity.

\subsubsection{Loss Function.} 

In the training phase, the overall objective function is:

\begin{equation}
    \mathcal{L}_{X\raisebox{0mm}{-}ReID} = \mathcal{L}_{Ins} + \lambda_1 \mathcal{L}_{IntraX} + \lambda_2 \mathcal{L}_{InterX}
\label{formula:total}
\end{equation}

\noindent where $\lambda_1$ and $\lambda_2$ are hyper-parameters to balance the entire loss. Notably, only the boosted Instance-Level features ${f}_{Ins}$ is used for inference. In this way, the final model has the same structure as Instance-Level Transformer whose input is a single image.

\section{Experiments}

\subsection{Evaluation Setting and Metrics}

Our proposed method are evaluated on the popular benchmark datasets: MSMT-17\cite{wei2018person} which contains 126441 person images of 4101 identities, Market1501\cite{zheng2015scalable} which contains 32668 person images of 1501 identities. Following the common setting, the cumulative matching characteristics (CMC) at Rank-1 and the mean average precision (mAP) are adopted to evaluate the performance of all methods. 

\subsection{Implementation Details}

We use the ImageNet\cite{deng2009imagenet}-pretrained model of ViT-Base, which contains $L=12$ Instance-Layers and the patch size is 16$\times$16. We build $M=N=12$ X-Layers in both IntraX and InterX. X-Layers in IntraX use an exponential moving average on Instance-Layer. X-layers are learnable in InterX. In Equation~\ref{formula:total}, we set $\lambda_1=5.0, \lambda_2=0.4$ and $\lambda_1=20.0, \lambda_2=0.4$ for Market1501 and MSMT17, respectively. The input images are resized to 256$\times$128. Following common practices, random horizontal flipping, padding, random cropping and random erasing\cite{zhong2020random} are used as augmentation for training. We randomly sample $K=4$ instances for each identity in a mini-batch with batch size 64. SGD optimizer with a momentum of 0.9 and the weight decay of 1e-4 is applied with a learning rate of 0.008 with cosine learning rate decay. We train 120 epochs in total. All the experiments are performed with one NVIDIA Geforce RTX 3090.

\subsection{Comparisons with State-of-the-art Methods}

In Table~\ref{table:sota}, we compare our proposed X-ReID with state-of-the-art methods on MSMT17 and Market1501. On Market1501, our method achieves comparable performance with the recent Transformer-based methods, especially on mAP. On the challenging MSMT17, Transformer-based methods outperform CNN-based methods significantly. X-ReID reaches state-of-the-art performance, which gains 1.1$\%$ mAP improvements when compared to the second place.

Our Baseline follows the setup in \cite{he2021transreid}. The Baselines of X-ReID and TransReID have the same implementation. In order to compare the experiments better, we reproduce Baseline and TransReID, whose performance is slightly different from that in the original paper due to different hardware environments. Our X-ReID is the Baseline with proposed IntraX and InterX. Based on the same Baseline, X-ReID outperforms TransReID by 1.6$\%$ mAP on MSMT17 and 0.6$\%$ mAP on Market1501.

% \usepackage{multirow}
\begin{table*}[t]
\caption{Comparison with state-of-the-art training methods. $\ast$ means we re-implement the experiments based on the official code. $\rm Baseline^{\ast}$ is Instance-Level Transformer described in Sec~\ref{sec:instance}, which follows the setup in \cite{he2021transreid}. $\rm TransReID^{\ast}$ is the result without side information for fair comparison.}
\begin{center}
% \scalebox{1.0}{
\begin{tabular}{c|c|c|cccc}
\hline
\multirow{2}{*}{Backbone}  &  \multirow{2}{*}{Methods}  &  \multirow{2}{*}{Ref}  & \multicolumn{2}{c|}{MSMT17}        & \multicolumn{2}{c}{Market1501}   \\ \cline{4-7} 
                      &             &       & mAP  & \multicolumn{1}{c|}{Rank-1} & mAP      & Rank-1        \\ \hline
\multirow{9}{*}{CNN}     & DuATM\cite{si2018dual}              & CVPR2018 & -           & \multicolumn{1}{c|}{-}             & 76.6          & 91.4         \\  
& GCSL\cite{chen2018group}              & CVPR2018 & -           & \multicolumn{1}{c|}{-}             & 81.6          & 93.5             \\
 & CASN\cite{zheng2019re}              & CVPR2019 & -           & \multicolumn{1}{c|}{-}             & 82.8          & 94.4     \\ 
                              & SFT\cite{luo2019spectral}               & ICCV2019 & 47.6        & \multicolumn{1}{c|}{73.6}          & 82.7          & 93.4         \\ \cline{2-7}
                              & PAT\cite{li2021diverse}               & CVPR2021 & -           & \multicolumn{1}{c|}{-}             & 88.0          & 95.4      \\
                              & ISP\cite{zhu2020identity}               & ECCV2020 & -           & \multicolumn{1}{c|}{-}             & \textbf{88.6}          & 95.3   \\
                              & PCB\cite{sun2018beyond}               & ECCV2018 & 40.4        & \multicolumn{1}{c|}{68.2}          & 81.6          & 93.8       \\
%                              & GASM\cite{he2020guided}              & ECCV2020 & 52.5        & \multicolumn{1}{c|}{79.5}          & 84.7          & \multicolumn{1}{c|}{95.3}            & 74.4         & 88.3          \\
                              & SCSN\cite{chen2020salience}              & CVPR2020 & 58.5        & \multicolumn{1}{c|}{\textbf{83.8}}          & 88.5          & {\textbf{95.7}}        \\
                              & ABDNet\cite{chen2019abd}            & ICCV2019 & \textbf{60.8}        & \multicolumn{1}{c|}{82.3}          & 88.3          & 95.6      \\ \hline \hline
\multirow{5}{*}{Transformer}     & $\rm Baseline^{\ast}$\cite{he2021transreid}          & ICCV2021 & 62.3 & \multicolumn{1}{c|}{82.2}   & 87.1 & 94.6       \\
% \multirow{5}{*}{\begin{tabular}[c]{@{}c@{}}Transformer\\ (ViT-Base)\end{tabular}}     & $\rm Baseline^{\ast}$\cite{he2021transreid}          & ICCV2021 & 62.3 & \multicolumn{1}{c|}{82.2}   & 87.1 & 94.6       \\
                              & AAformer\cite{zhu2021aaformer}         & Arxiv2021 & 63.2 & \multicolumn{1}{c|}{83.6}   & 87.7 & {\textbf{95.2}}       \\
                              & $\rm TransReID^{\ast}$\cite{he2021transreid}         & ICCV2021 & 63.5 & \multicolumn{1}{c|}{83.0}   & 87.4 & 94.2      \\
                              & DCAL\cite{zhu2022dual}         & CVPR2022 & 64.0 & \multicolumn{1}{c|}{83.1}   & 87.5 & 94.7      \\
                              & X-ReID           & Ours     & \textbf{65.1} & \multicolumn{1}{c|}{\textbf{84.0}}   & \textbf{88.0} & 94.9  \\ \hline
%                              & X-ReID(AttnRank) & Ours     & 75.3 & \multicolumn{1}{c|}{86.3}   & 92.3 & \multicolumn{1}{c|}{95.3}   & 87.1         & 91.4          \\ \hline
% \multirow{3}{*}{ViT-B/16+SIE\cite{he2021transreid}} & $\rm Baseline^{\ast}$\cite{he2021transreid}          & ICCV2021 & 63.9 & \multicolumn{1}{c|}{82.4}   & 88.1 & \multicolumn{1}{c|}{94.5}   & 80.8         & 89.5          \\
%                               & $\rm TransReID^{\ast}$\cite{he2021transreid}         & ICCV2021 & 65.1 & \multicolumn{1}{c|}{83.2}   & 88.2 & \multicolumn{1}{c|}{\textbf{95.1}}   & 80.7         & 89.5          \\
%                               & X-ReID           & Ours     & \textbf{66.5} & \multicolumn{1}{c|}{\textbf{84.3}}   & \textbf{88.6} & \multicolumn{1}{c|}{94.9}   & \textbf{81.7}         & \textbf{90.0}          \\ \hline
%                              & X-ReID(AttnRank) & Ours     & 76.9 & \multicolumn{1}{c|}{86.9}   & 92.7 & \multicolumn{1}{c|}{95.7}   & 87.4         & 91.3          \\ \hline
\end{tabular}
% }

\end{center}
\label{table:sota}
\end{table*}

CNN-based methods in the 1st group leverage the cross-image information, which mine the relations among Instance-Level features. Transformer-based DCAL\cite{zhu2022dual} proposes cross-attention to learn local features better, which is classified as a Part-Level method. However, our proposed Identity-Level method are introduced to extract more unified features among all intra-identity instances and push inter-identity instances far away. X-ReID surpasses the CNN-based cross-image methods by a large margin, at least 5.3$\%$ mAP on Market1501. X-ReID outperforms the cross-attention related method DCAL by 1.1$\%$ mAP on MSMT17 and 0.5$\%$ mAP on Market1501.

\subsection{Effectiveness of X-ReID}

We propose to improve Instance-Level features with our Identity-Level framework. Compared to Instance-Level features generated from a single image, Identity-Level features, generated across different images of each identity, have more unified pedestrian information. So that in X-ReID, different Instance-Level features of each identity are very compact and features from different identities are easy to distinguish, which facilitates the model to learn more discriminative and unified features.

\subsubsection{Compactness Analysis.}
\label{sec:compact}

%To investigate the compactness of Instance-Level features improved by X-ReID, we evaluate the metric of compactness (CP), which is formulated as:
We evaluate intra-identity variation as follows:
%The compactness is used to evaluate within-identity variations:

\begin{equation}
\begin{split}
\omega_i=\frac{1}{|\Omega_i|}\sum_{x_j\in\Omega_i}x_j,CP_i & =\frac{1}{|\Omega_i|}\sum_{x_j\in\Omega_i}\left\| x_j - \omega_i \right\| \\ CP & =\frac{1}{N}\sum_{k=1}^{N}CP_k  
\end{split}
\end{equation}

\begin{figure}
  \begin{subfigure}{.49\linewidth}
    \centering
    \includegraphics[width=1.0\linewidth]{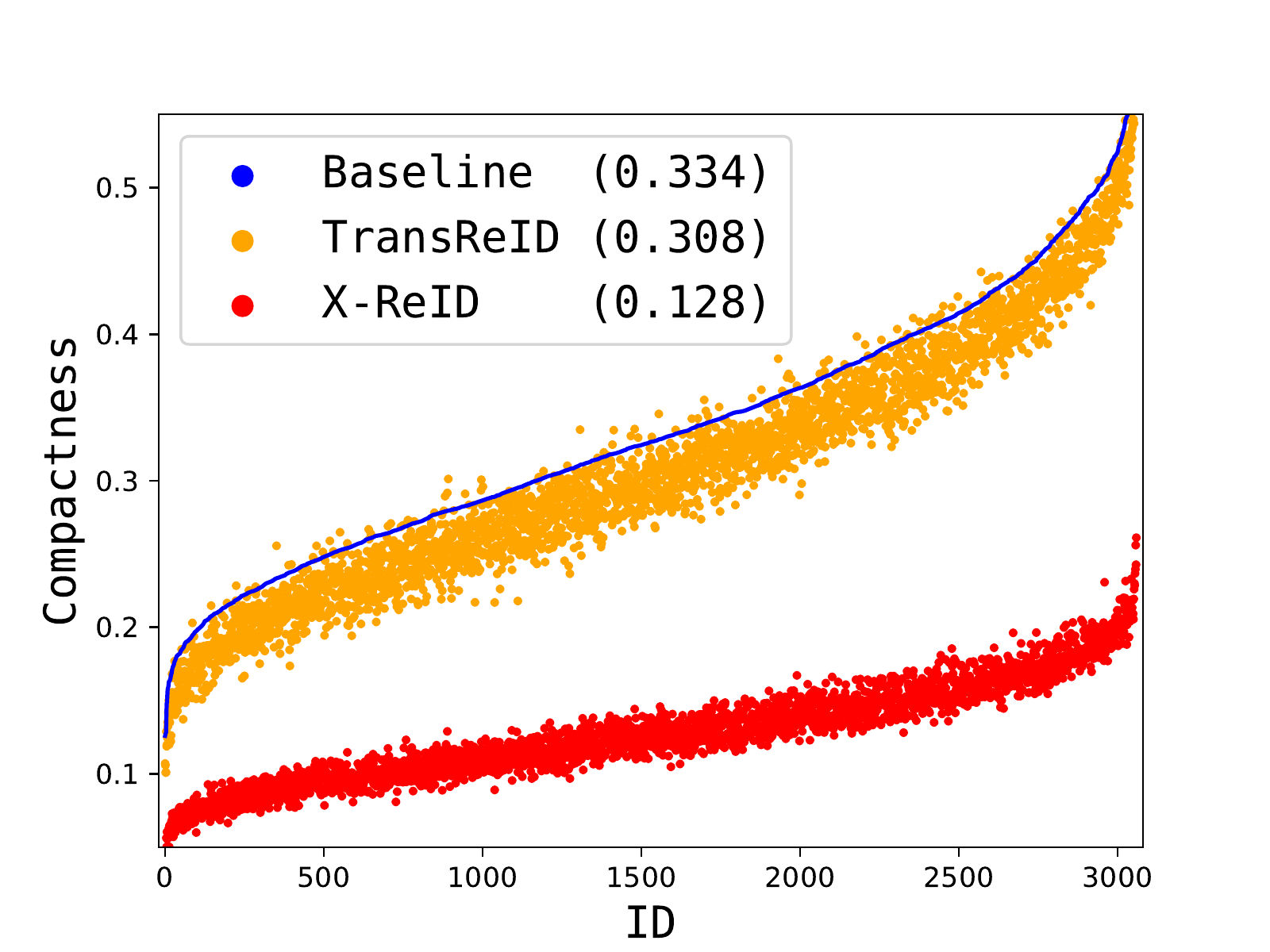}
    \caption{MSMT17}
  \end{subfigure}
  \begin{subfigure}{.49\linewidth}
    \centering
    \includegraphics[width=1.0\linewidth]{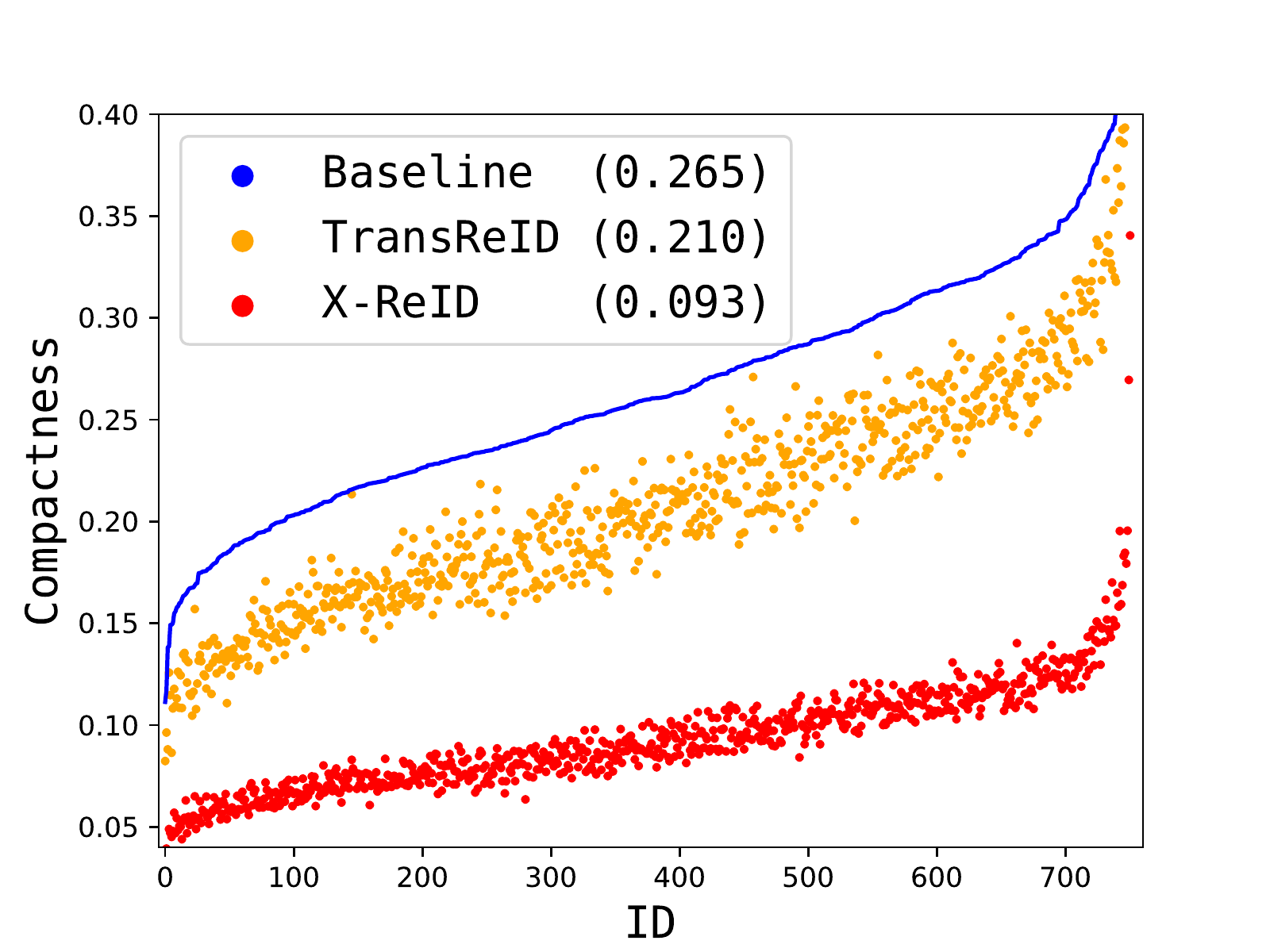}
    \caption{Market1501}
  \end{subfigure}
  \caption{Compactness analysis in the test set on MSMT17 and Market1501. The order of the person ID axis is in ascending order of the compactness of Baseline. The overall compactness is shown in the parentheses.}
  \label{fig:cp}
\end{figure}

\noindent where $\Omega_i$ denotes the set containing all the feature vectors of the $i$-th person and $\omega_i$ represents the feature center of the $i$-th person. $CP_i$ calculates the average distance from each instance to the center in the $i$-th person. $N$ is the number of persons and the overall compactness $CP$ is the mean compactness of all persons. The lower the value of $CP$, the more compact the features of each identity. Figure~\ref{fig:cp} shows the compactness of different persons in the test set on MSMT17 and Market1501. Although TransReID outperforms Baseline, the compactness is almost the same. With Identity-Level, the compactness is much lower than Baseline and TransReID. Such results prove the Identity-Level knowledge can reduce intra-identity variation.
% improve the compactness of Instance-Level features

\subsubsection{The Calinski-Harabasz Score Analysis.}

\begin{figure}
  \begin{subfigure}{.49\linewidth}
    \centering
    \includegraphics[width=1.0\linewidth]{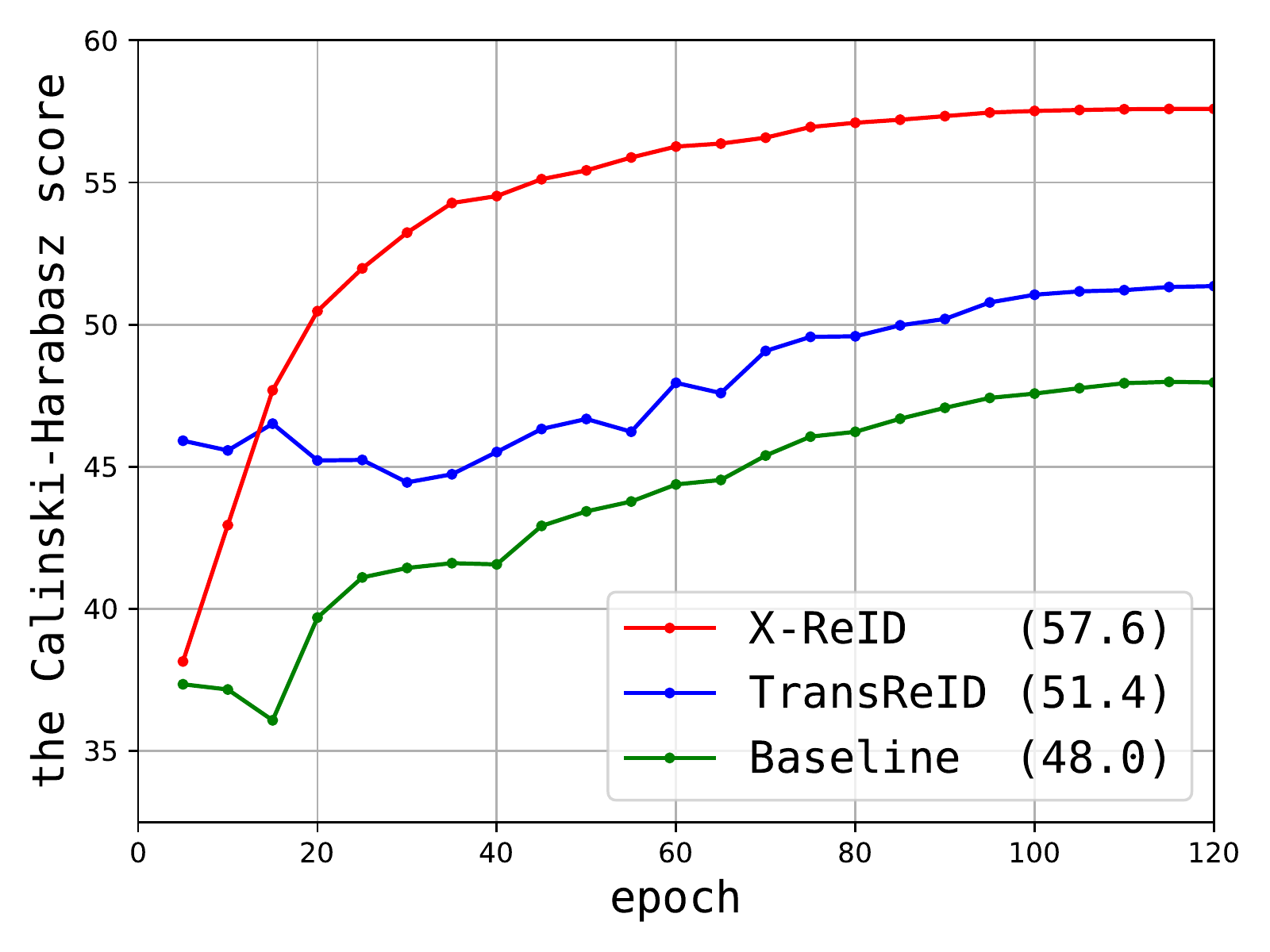}
    \caption{MSMT17}
  \end{subfigure}
  \begin{subfigure}{.49\linewidth}
    \centering
    \includegraphics[width=1.0\linewidth]{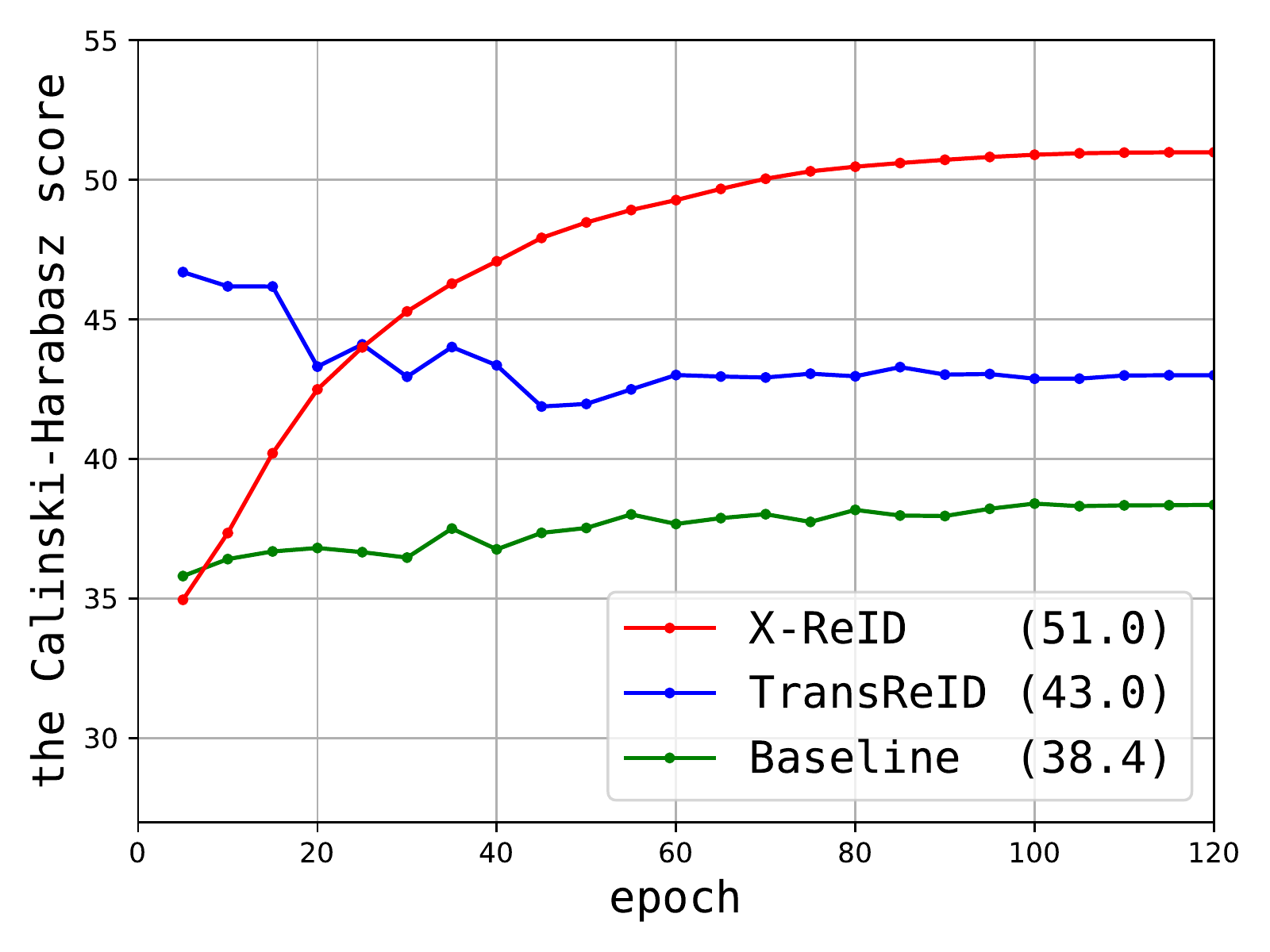}
    \caption{Market1501}
  \end{subfigure}
  \caption{The \textbf{C}alinski-\textbf{H}arabasz score analysis in the test set on MSMT17 and Market1501. CH is calculated every 5 training epochs. The value in the parentheses is CH of the final model.}
  \label{fig:ch}
\end{figure}

The \textbf{C}alinski-\textbf{H}arabasz score~\cite{calinski1974dendrite} is defined as ratio of the sum of between-cluster dispersion and of within-cluster dispersion, which can be used to evaluate intra- and inter-identity variations. CH is higher when identities are both dense and well separated. Fig.\ref{fig:ch} shows CH metric during the training epochs. We can find that the curves of TransReID and Baseline rise slowly and the curve of X-ReID rises rapidly. The curve of X-ReID is significantly higher than that of TransReID and Baseline. By our proposed method, intra-identity instances are more compact and inter-identity instances are well separated. 

\begin{figure}[t]
\centering
\scalebox{0.85}
{
\includegraphics[height=\linewidth, angle=-90]{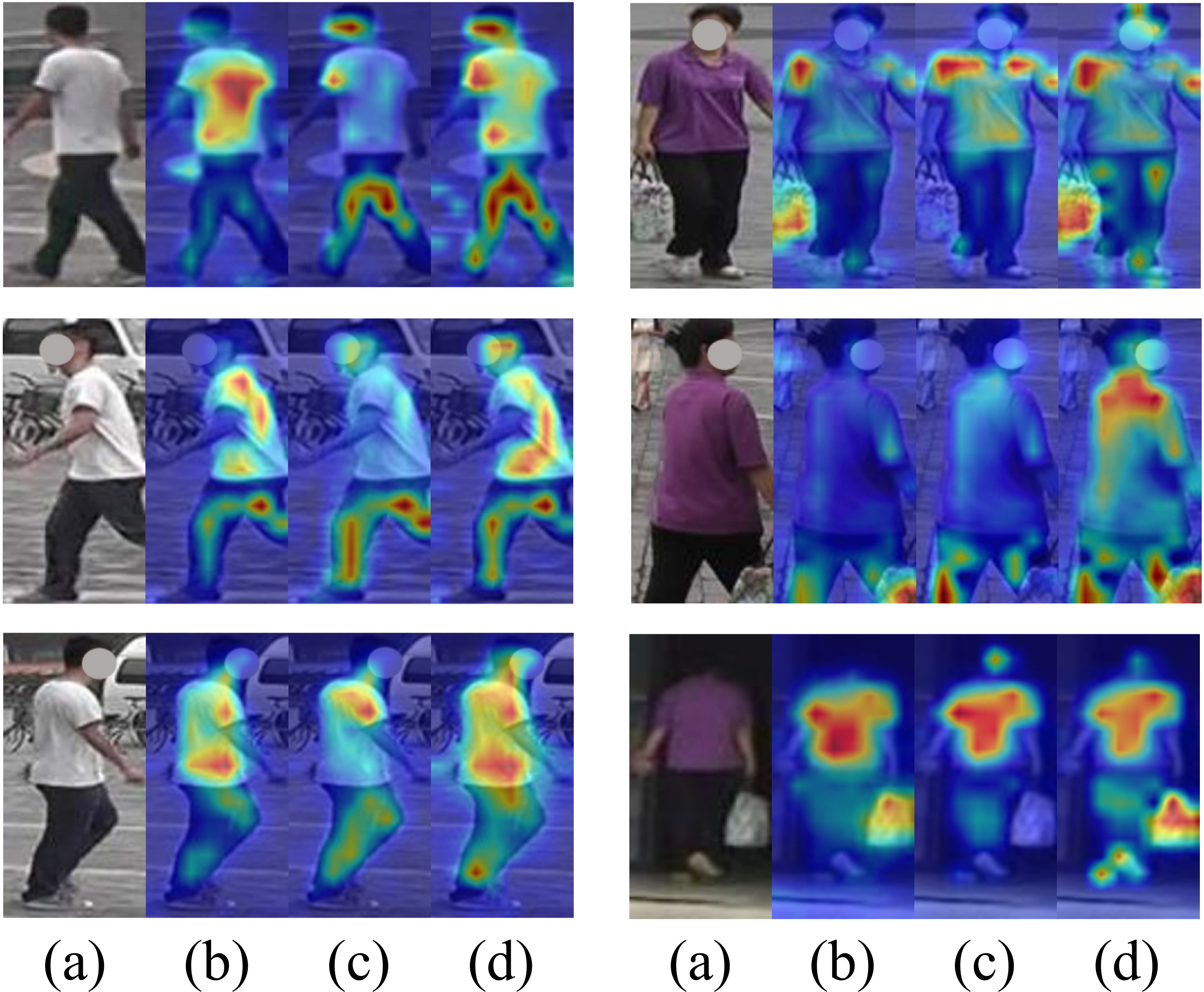}
}
\caption{Visualization analysis of attention maps. (a) Input images, (b) Baseline, (c) TransReID, (d) X-ReID. The images in each column are the same identity.}
\label{fig:vis}
\end{figure}

\subsubsection{Visualization Analysis.}

In  Figure~\ref{fig:vis}, we further conduct visualization experiments\cite{chefer2021transformer} to show the effectiveness of X-ReID. We compare our method with Baseline and TransReID. Our model focuses on more comprehensive areas, while high attention regions are relatively scattered instead of focusing on a small area, which indicates that our model can obtain more unified pedestrian information. In the second column, the ``handbag'' is only a small part of the middle image. Our model learns the Identity-Level information from multiple images to improve the discriminative ability to recognize the ``handbag''. 

\subsection{Ablation Studies of X-ReID}

\subsubsection{The effectiveness of individual components.}

There are two important components in our proposed X-ReID: a Cross Intra-Identity Instances module (IntraX) and a Cross Inter-Identity Instances module (InterX). We evaluate the contribution of each component in Table~\ref{table:ablation}. EMA means the \textbf{E}xponential \textbf{M}oving \textbf{A}verage operation in IntraX. The improvement of Baseline+IntraX is more impressive than that of Baseline+EMA, which validates the effectiveness of Identity-Level rather than EMA. Through fusing intra-identity instances, IntraX transfer the Identity-Level knowledge to Instance-Level features to gain promising performance. InterX deals with inter-identity instances to minimize intra-identity variation and maximize inter-identity variation. Compared with Baseline, our full X-ReID significantly boosts the performance by 2.8$\%$ mAP and 0.9$\%$ mAP on MSMT17 and Market1501, respectively.

\begin{table}[t]
\centering
\caption{The ablation study on individual components of X-ReID. EMA is IntraX without $\mathcal{L}_{IntraX}$.}
\begin{tabular}{c|cc|cc}
\hline
\multirow{2}{*}{Methods} & \multicolumn{2}{c|}{MSMT17} & \multicolumn{2}{c}{Market1501} \\ \cline{2-5} 
                        & mAP         & Rank-1        & mAP           & Rank-1         \\ \hline
Baseline                & 62.3        & 82.2          & 87.1          & 94.6           \\
Baseline+EMA            & 62.4        & 82.3          & 87.1          & 94.7           \\
Baseline+IntraX         & 64.7        & 83.4          & 87.8          & 94.7           \\
Baseline+InterX         & 63.7        & 83.2          & 87.4          & 94.7           \\
X-ReID                  & 65.1        & 84.0          & 88.0          & 94.9           \\ \hline
\end{tabular}
\label{table:ablation}
\end{table}

% \begin{figure}[t]
% \centering
% \includegraphics[height=\linewidth, angle=-90]{Fig/attention.pdf}
% \caption{Parameters analysis of hyper-parameters $\lambda_1$ and $\lambda_2$ of the entire loss on MSMT17 and Market1501. The first row shows the impact of $\lambda_1$ when $\lambda_2$ is fixed at 0.4 and the second row shows the impact of $\lambda_2$ when $\lambda_1$ is fixed.}
% \label{fig:Hyper}
% \end{figure}

\begin{figure}[t]
  \begin{subfigure}{.49\linewidth}
    \centering
    \includegraphics[width=1.0\linewidth]{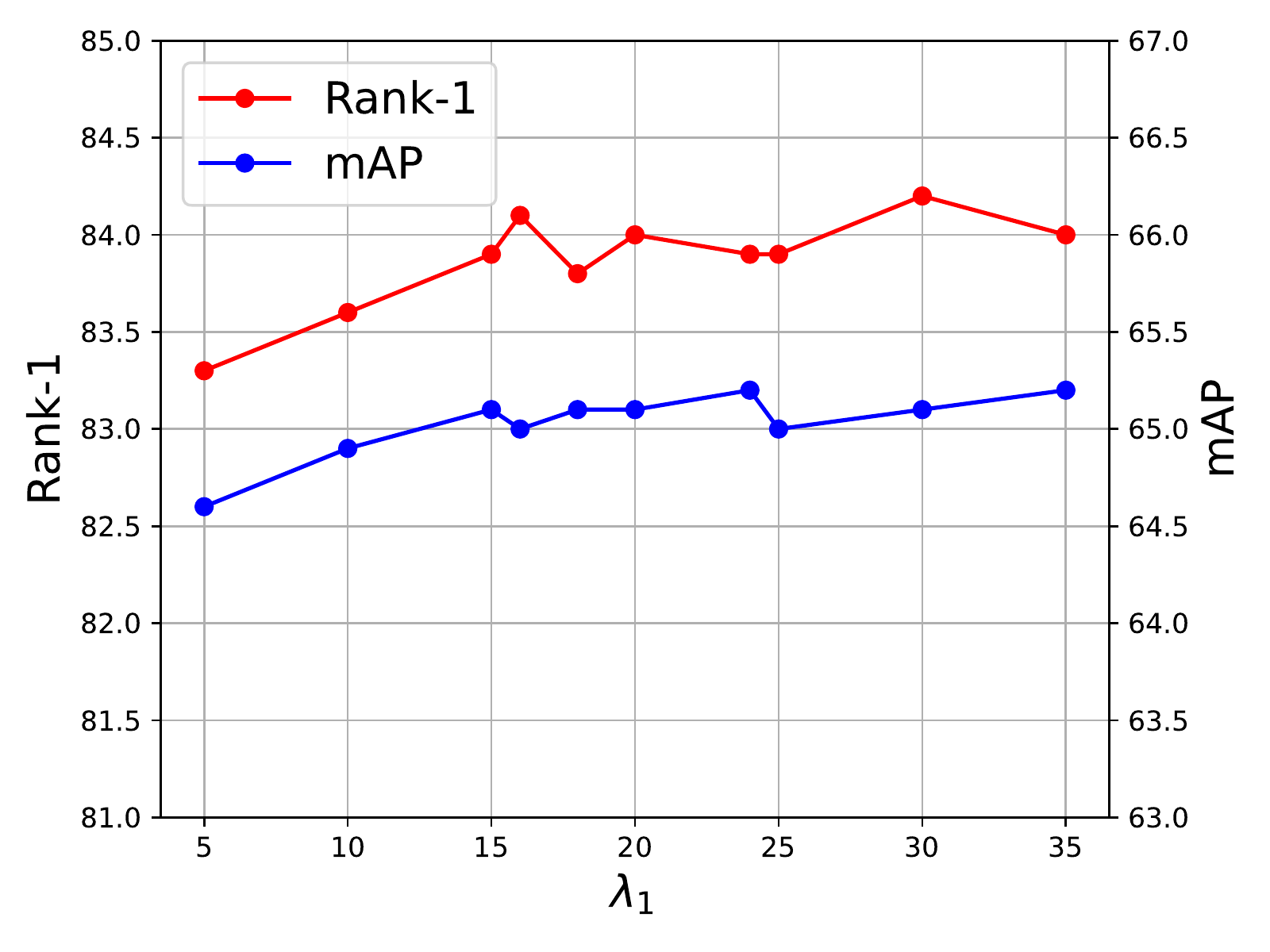}
    \caption{MSMT17}
  \end{subfigure}
  \begin{subfigure}{.49\linewidth}
    \centering
    \includegraphics[width=1.0\linewidth]{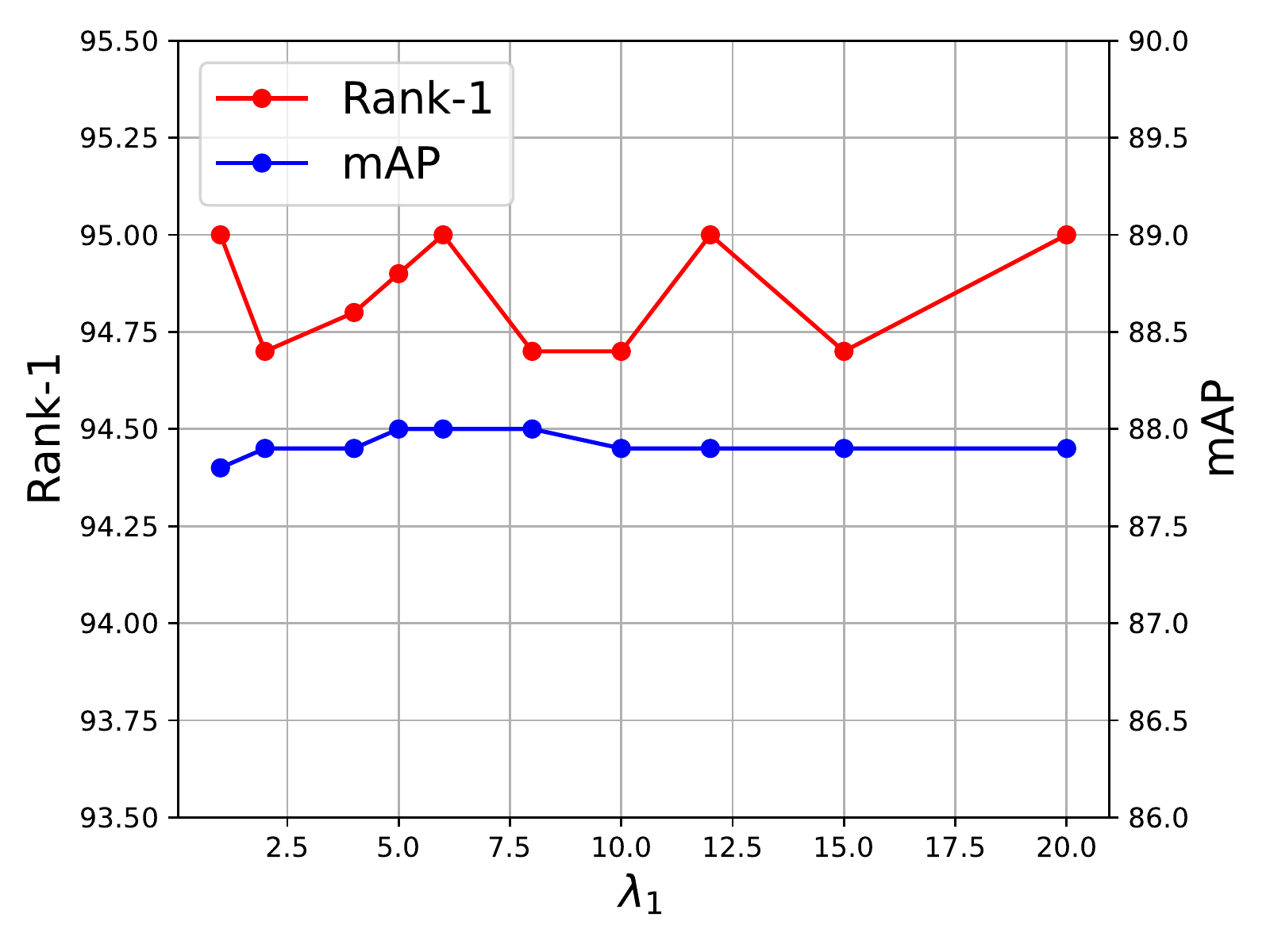}
    \caption{Market1501}
  \end{subfigure}

  \begin{subfigure}{.49\linewidth}
    \centering
    \includegraphics[width=1.0\linewidth]{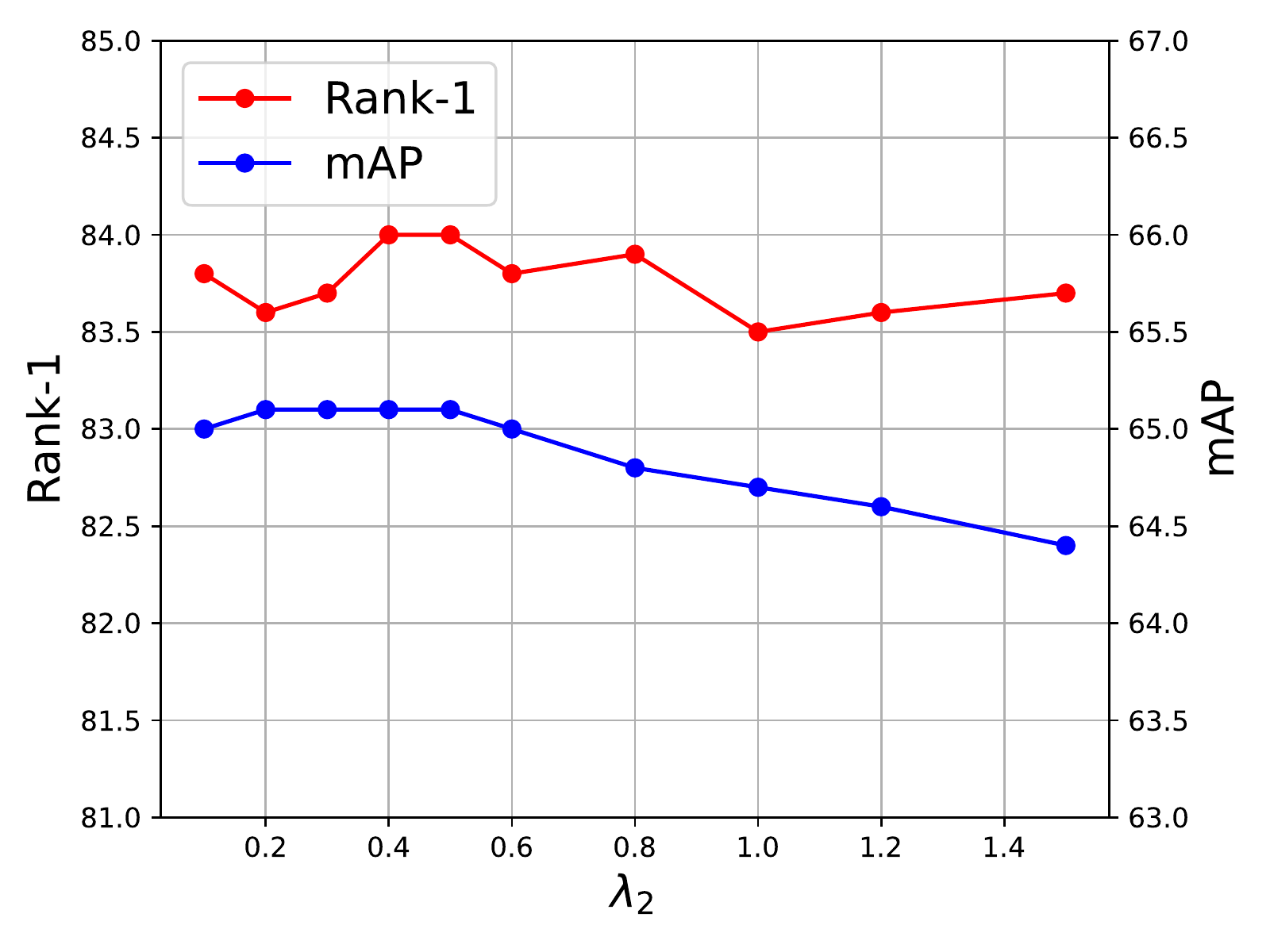}
    \caption{MSMT17}
  \end{subfigure}
  \begin{subfigure}{.49\linewidth}
    \centering
    \includegraphics[width=1.0\linewidth]{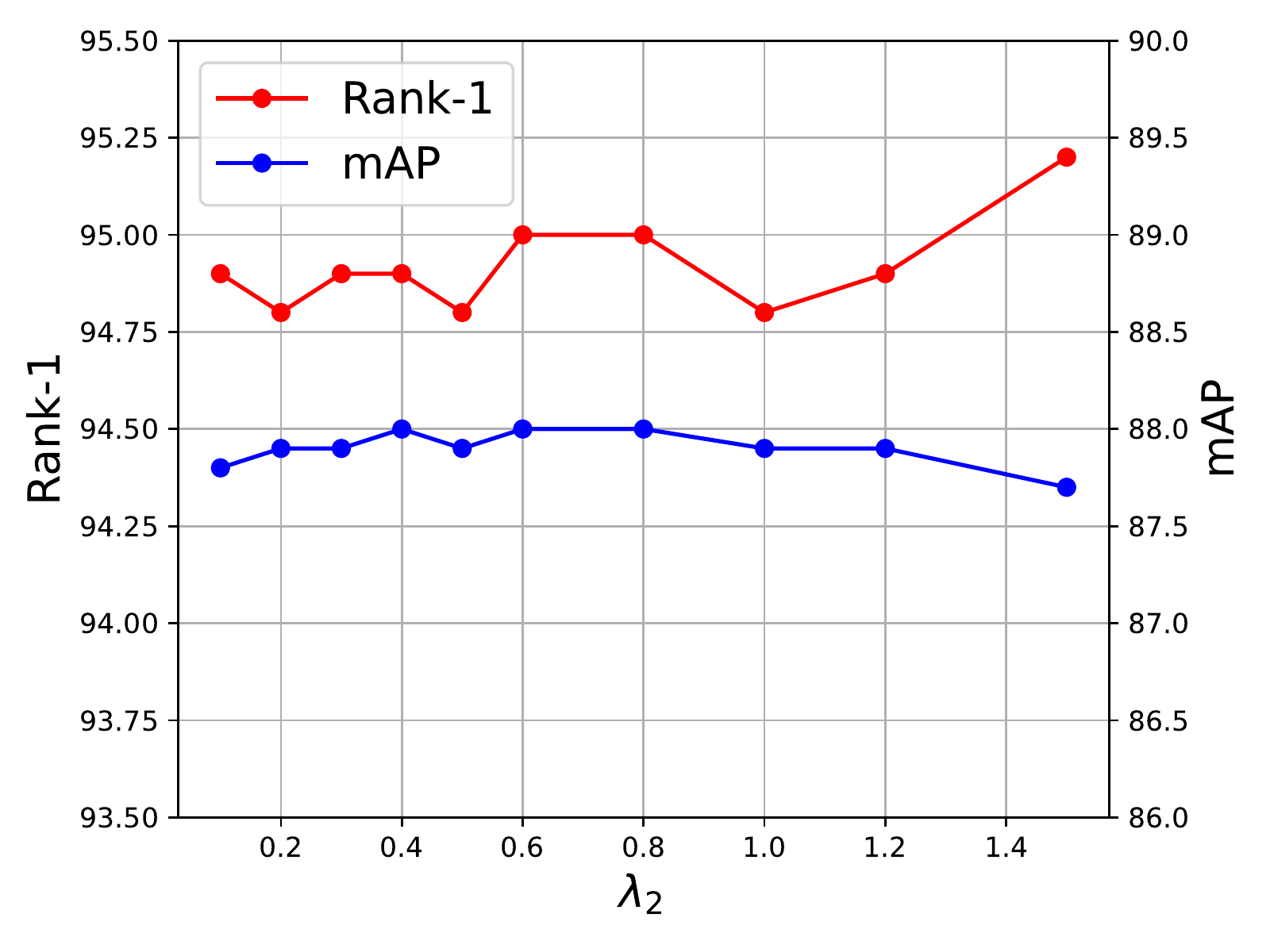}
    \caption{Market1501}
  \end{subfigure}
  
  \caption{Hyper-parameters analysis of weight $\lambda_1$ and $\lambda_2$ of the entire loss on MSMT17 and Market1501.}
  \label{fig:lambda}
\end{figure}
% The first row shows the impact of $\lambda_1$ when $\lambda_2$ is fixed at 0.4 and the second row shows the impact of $\lambda_2$ when $\lambda_1$ is fixed.

\subsubsection{Hyper-parameters Analysis}

We evaluate the impact of weight $\lambda_1$ and $\lambda_2$ of the entire loss (Equation~\ref{formula:total}) in Figure~\ref{fig:lambda}. $\lambda_1$ balances the weight of IntraX and $\lambda_2$ balances the weight of InterX. The first row shows the impact of $\lambda_1$ when $\lambda_2$ is fixed at 0.4. Although the final selected $\lambda_1$ is not the same on the two datasets, the performance fluctuates very little with the change of $\lambda_1$ and mAP is at least above 64.6$\%$ on MSMT17. The second row shows the impact of $\lambda_2$ when $\lambda_1$ is fixed at 20.0 on MSMT17 and 5.0 on Market1501. The performance obtains a further improvement when the value of $\lambda_2$ is around 0.4. 

% These experiments validate that our method is insensitive to $\lambda_1$ and $\lambda_2$.

% % Please add the following required packages to your document preamble:
% % \usepackage{multirow}
% \begin{table}[]
% \centering
% \caption{Ablation study of layers of IntraX and InterX.}
% \begin{tabular}{cc|cc|cc}
% \hline
% \multirow{2}{*}{M} & \multirow{2}{*}{N} & \multicolumn{2}{c|}{MSMT17} & \multicolumn{2}{c}{Market1501} \\ \cline{3-6} 
%                    &                    & mAP         & Rank-1        & mAP           & Rank-1         \\ \hline
% 12                 & 4                  & 65.0        & 83.8          & 87.9          & 94.7           \\
% 12                 & 7                  & 65.0        & 83.7          & 87.9          & 94.7           \\
% 12                 & 10                 & 65.0        & 83.6          & 87.9          & 94.7           \\ \hline \hline
% 4                  & 12                 & 65.1        & 83.9          & 87.9          & 94.9           \\
% 7                  & 12                 & 65.1        & 83.9          & 88.0          & 94.9           \\
% 10                 & 12                 & 65.1        & 83.8          & 88.0          & 94.9           \\ \hline \hline
% 12                 & 12                 & 65.1        & 84.0          & 88.0          & 94.9           \\ \hline
% \end{tabular}
% \end{table}

\section{Conclusion and Future Work}

In this paper, we propose a novel training framework named X-ReID, which contains a Cross Intra-Identity Instances module and a Cross Inter-Identity Instances module. Our proposed methods extract Identity-Level features for more unified and discriminative pedestrian information. Extensive experiments on benchmark datasets show the superiority of our methods over state-of-the-art methods. In the future, more sophisticated solutions for improving the Identity-Level capability can be exploited.

\clearpage
\bibliography{aaai23}

\begin{thebibliography}{44}
\providecommand{\natexlab}[1]{#1}

\bibitem[{Ba, Kiros, and Hinton(2016)}]{ba2016layer}
Ba, J.~L.; Kiros, J.~R.; and Hinton, G.~E. 2016.
\newblock Layer normalization.
\newblock \emph{arXiv preprint arXiv:1607.06450}.

\bibitem[{Cali{\'n}ski and Harabasz(1974)}]{calinski1974dendrite}
Cali{\'n}ski, T.; and Harabasz, J. 1974.
\newblock A dendrite method for cluster analysis.
\newblock \emph{Communications in Statistics-theory and Methods}, 3(1): 1--27.

\bibitem[{Carion et~al.(2020)Carion, Massa, Synnaeve, Usunier, Kirillov, and
  Zagoruyko}]{carion2020end}
Carion, N.; Massa, F.; Synnaeve, G.; Usunier, N.; Kirillov, A.; and Zagoruyko,
  S. 2020.
\newblock End-to-end object detection with transformers.
\newblock In \emph{European conference on computer vision}, 213--229. Springer.

\bibitem[{Chefer, Gur, and Wolf(2021)}]{chefer2021transformer}
Chefer, H.; Gur, S.; and Wolf, L. 2021.
\newblock Transformer interpretability beyond attention visualization.
\newblock In \emph{Proceedings of the IEEE/CVF Conference on Computer Vision
  and Pattern Recognition}, 782--791.

\bibitem[{Chen, Fan, and Panda(2021)}]{chen2021crossvit}
Chen, C.-F.~R.; Fan, Q.; and Panda, R. 2021.
\newblock Crossvit: Cross-attention multi-scale vision transformer for image
  classification.
\newblock In \emph{Proceedings of the IEEE/CVF International Conference on
  Computer Vision}, 357--366.

\bibitem[{Chen et~al.(2018)Chen, Xu, Li, Sebe, and Wang}]{chen2018group}
Chen, D.; Xu, D.; Li, H.; Sebe, N.; and Wang, X. 2018.
\newblock Group consistent similarity learning via deep crf for person
  re-identification.
\newblock In \emph{Proceedings of the IEEE conference on computer vision and
  pattern recognition}, 8649--8658.

\bibitem[{Chen et~al.(2021)Chen, Wang, Guo, Xu, Deng, Liu, Ma, Xu, Xu, and
  Gao}]{chen2021pre}
Chen, H.; Wang, Y.; Guo, T.; Xu, C.; Deng, Y.; Liu, Z.; Ma, S.; Xu, C.; Xu, C.;
  and Gao, W. 2021.
\newblock Pre-trained image processing transformer.
\newblock In \emph{Proceedings of the IEEE/CVF Conference on Computer Vision
  and Pattern Recognition}, 12299--12310.

\bibitem[{Chen et~al.(2019)Chen, Ding, Xie, Yuan, Chen, Yang, Ren, and
  Wang}]{chen2019abd}
Chen, T.; Ding, S.; Xie, J.; Yuan, Y.; Chen, W.; Yang, Y.; Ren, Z.; and Wang,
  Z. 2019.
\newblock Abd-net: Attentive but diverse person re-identification.
\newblock In \emph{Proceedings of the IEEE/CVF International Conference on
  Computer Vision}, 8351--8361.

\bibitem[{Chen et~al.(2020)Chen, Fu, Zhao, Zheng, Song, Ji, and
  Yang}]{chen2020salience}
Chen, X.; Fu, C.; Zhao, Y.; Zheng, F.; Song, J.; Ji, R.; and Yang, Y. 2020.
\newblock Salience-guided cascaded suppression network for person
  re-identification.
\newblock In \emph{Proceedings of the IEEE/CVF Conference on Computer Vision
  and Pattern Recognition}, 3300--3310.

\bibitem[{Deng et~al.(2009)Deng, Dong, Socher, Li, Li, and
  Fei-Fei}]{deng2009imagenet}
Deng, J.; Dong, W.; Socher, R.; Li, L.-J.; Li, K.; and Fei-Fei, L. 2009.
\newblock Imagenet: A large-scale hierarchical image database.
\newblock In \emph{2009 IEEE conference on computer vision and pattern
  recognition}, 248--255. Ieee.

\bibitem[{Dosovitskiy et~al.(2020)Dosovitskiy, Beyer, Kolesnikov, Weissenborn,
  Zhai, Unterthiner, Dehghani, Minderer, Heigold, Gelly
  et~al.}]{dosovitskiy2020image}
Dosovitskiy, A.; Beyer, L.; Kolesnikov, A.; Weissenborn, D.; Zhai, X.;
  Unterthiner, T.; Dehghani, M.; Minderer, M.; Heigold, G.; Gelly, S.; et~al.
  2020.
\newblock An image is worth 16x16 words: Transformers for image recognition at
  scale.
\newblock \emph{arXiv preprint arXiv:2010.11929}.

\bibitem[{Fang et~al.(2019)Fang, Zhou, Roy, Petersson, and
  Harandi}]{fang2019bilinear}
Fang, P.; Zhou, J.; Roy, S.~K.; Petersson, L.; and Harandi, M. 2019.
\newblock Bilinear attention networks for person retrieval.
\newblock In \emph{Proceedings of the IEEE/CVF international conference on
  computer vision}, 8030--8039.

\bibitem[{Gao et~al.(2021)Gao, Wan, Pan, Peng, Tian, Han, Zhou, and
  Ye}]{gao2021ts}
Gao, W.; Wan, F.; Pan, X.; Peng, Z.; Tian, Q.; Han, Z.; Zhou, B.; and Ye, Q.
  2021.
\newblock TS-CAM: Token Semantic Coupled Attention Map for Weakly Supervised
  Object Localization.
\newblock In \emph{Proceedings of the IEEE/CVF International Conference on
  Computer Vision}, 2886--2895.

\bibitem[{Han et~al.(2020)Han, Wang, Chen, Chen, Guo, Liu, Tang, Xiao, Xu, Xu
  et~al.}]{han2020survey}
Han, K.; Wang, Y.; Chen, H.; Chen, X.; Guo, J.; Liu, Z.; Tang, Y.; Xiao, A.;
  Xu, C.; Xu, Y.; et~al. 2020.
\newblock A survey on visual transformer.
\newblock \emph{arXiv e-prints}, arXiv--2012.

\bibitem[{He et~al.(2016)He, Zhang, Ren, and Sun}]{he2016deep}
He, K.; Zhang, X.; Ren, S.; and Sun, J. 2016.
\newblock Deep residual learning for image recognition.
\newblock In \emph{Proceedings of the IEEE conference on computer vision and
  pattern recognition}, 770--778.

\bibitem[{He and Liu(2020)}]{he2020guided}
He, L.; and Liu, W. 2020.
\newblock Guided saliency feature learning for person re-identification in
  crowded scenes.
\newblock In \emph{European Conference on Computer Vision}, 357--373. Springer.

\bibitem[{He et~al.(2021)He, Luo, Wang, Wang, Li, and Jiang}]{he2021transreid}
He, S.; Luo, H.; Wang, P.; Wang, F.; Li, H.; and Jiang, W. 2021.
\newblock Transreid: Transformer-based object re-identification.
\newblock In \emph{Proceedings of the IEEE/CVF International Conference on
  Computer Vision}, 15013--15022.

\bibitem[{Hermans, Beyer, and Leibe(2017)}]{hermans2017defense}
Hermans, A.; Beyer, L.; and Leibe, B. 2017.
\newblock In defense of the triplet loss for person re-identification.
\newblock \emph{arXiv preprint arXiv:1703.07737}.

\bibitem[{Jin et~al.(2020)Jin, Lan, Zeng, Wei, and Chen}]{jin2020semantics}
Jin, X.; Lan, C.; Zeng, W.; Wei, G.; and Chen, Z. 2020.
\newblock Semantics-aligned representation learning for person
  re-identification.
\newblock In \emph{Proceedings of the AAAI Conference on Artificial
  Intelligence}, volume~34, 11173--11180.

\bibitem[{Khan et~al.(2021)Khan, Naseer, Hayat, Zamir, Khan, and
  Shah}]{khan2021transformers}
Khan, S.; Naseer, M.; Hayat, M.; Zamir, S.~W.; Khan, F.~S.; and Shah, M. 2021.
\newblock Transformers in vision: A survey.
\newblock \emph{ACM Computing Surveys (CSUR)}.

\bibitem[{Li et~al.(2021{\natexlab{a}})Li, Hou, Wang, Gao, Xu, and
  Li}]{li2021trear}
Li, X.; Hou, Y.; Wang, P.; Gao, Z.; Xu, M.; and Li, W. 2021{\natexlab{a}}.
\newblock Trear: Transformer-based rgb-d egocentric action recognition.
\newblock \emph{IEEE Transactions on Cognitive and Developmental Systems}.

\bibitem[{Li et~al.(2021{\natexlab{b}})Li, He, Zhang, Liu, Zhang, and
  Wu}]{li2021diverse}
Li, Y.; He, J.; Zhang, T.; Liu, X.; Zhang, Y.; and Wu, F. 2021{\natexlab{b}}.
\newblock Diverse part discovery: Occluded person re-identification with
  part-aware transformer.
\newblock In \emph{Proceedings of the IEEE/CVF Conference on Computer Vision
  and Pattern Recognition}, 2898--2907.

\bibitem[{Liu et~al.(2017)Liu, Feng, Qi, Jiang, and Yan}]{liu2017end}
Liu, H.; Feng, J.; Qi, M.; Jiang, J.; and Yan, S. 2017.
\newblock End-to-end comparative attention networks for person
  re-identification.
\newblock \emph{IEEE Transactions on Image Processing}, 26(7): 3492--3506.

\bibitem[{Luo et~al.(2019{\natexlab{a}})Luo, Chen, Wang, and
  Zhang}]{luo2019spectral}
Luo, C.; Chen, Y.; Wang, N.; and Zhang, Z. 2019{\natexlab{a}}.
\newblock Spectral feature transformation for person re-identification.
\newblock In \emph{Proceedings of the IEEE/CVF International Conference on
  Computer Vision}, 4976--4985.

\bibitem[{Luo et~al.(2019{\natexlab{b}})Luo, Gu, Liao, Lai, and
  Jiang}]{luo2019bag}
Luo, H.; Gu, Y.; Liao, X.; Lai, S.; and Jiang, W. 2019{\natexlab{b}}.
\newblock Bag of tricks and a strong baseline for deep person
  re-identification.
\newblock In \emph{Proceedings of the IEEE/CVF conference on computer vision
  and pattern recognition workshops}, 0--0.

\bibitem[{Si et~al.(2018)Si, Zhang, Li, Kuen, Kong, Kot, and Wang}]{si2018dual}
Si, J.; Zhang, H.; Li, C.-G.; Kuen, J.; Kong, X.; Kot, A.~C.; and Wang, G.
  2018.
\newblock Dual attention matching network for context-aware feature sequence
  based person re-identification.
\newblock In \emph{Proceedings of the IEEE conference on computer vision and
  pattern recognition}, 5363--5372.

\bibitem[{Sun et~al.(2018)Sun, Zheng, Yang, Tian, and Wang}]{sun2018beyond}
Sun, Y.; Zheng, L.; Yang, Y.; Tian, Q.; and Wang, S. 2018.
\newblock Beyond part models: Person retrieval with refined part pooling (and a
  strong convolutional baseline).
\newblock In \emph{Proceedings of the European conference on computer vision},
  480--496.

\bibitem[{Vaswani et~al.(2017)Vaswani, Shazeer, Parmar, Uszkoreit, Jones,
  Gomez, Kaiser, and Polosukhin}]{vaswani2017attention}
Vaswani, A.; Shazeer, N.; Parmar, N.; Uszkoreit, J.; Jones, L.; Gomez, A.~N.;
  Kaiser, {\L}.; and Polosukhin, I. 2017.
\newblock Attention is all you need.
\newblock \emph{Advances in neural information processing systems}, 30.

\bibitem[{Wang et~al.(2018)Wang, Yuan, Chen, Li, and Zhou}]{wang2018learning}
Wang, G.; Yuan, Y.; Chen, X.; Li, J.; and Zhou, X. 2018.
\newblock Learning discriminative features with multiple granularities for
  person re-identification.
\newblock In \emph{Proceedings of the 26th ACM international conference on
  Multimedia}, 274--282.

\bibitem[{Wang et~al.(2022)Wang, Shen, Liu, Gao, and Gavves}]{wang2022nformer}
Wang, H.; Shen, J.; Liu, Y.; Gao, Y.; and Gavves, E. 2022.
\newblock NFormer: Robust Person Re-identification with Neighbor Transformer.
\newblock In \emph{Proceedings of the IEEE/CVF Conference on Computer Vision
  and Pattern Recognition}, 7297--7307.

\bibitem[{Wei et~al.(2018)Wei, Zhang, Gao, and Tian}]{wei2018person}
Wei, L.; Zhang, S.; Gao, W.; and Tian, Q. 2018.
\newblock Person transfer gan to bridge domain gap for person
  re-identification.
\newblock In \emph{Proceedings of the IEEE conference on computer vision and
  pattern recognition}, 79--88.

\bibitem[{Xie et~al.(2021)Xie, Wang, Yu, Anandkumar, Alvarez, and
  Luo}]{xie2021segformer}
Xie, E.; Wang, W.; Yu, Z.; Anandkumar, A.; Alvarez, J.~M.; and Luo, P. 2021.
\newblock SegFormer: Simple and efficient design for semantic segmentation with
  transformers.
\newblock \emph{Advances in Neural Information Processing Systems}, 34.

\bibitem[{Ye et~al.(2021)Ye, Shen, Lin, Xiang, Shao, and Hoi}]{ye2021deep}
Ye, M.; Shen, J.; Lin, G.; Xiang, T.; Shao, L.; and Hoi, S.~C. 2021.
\newblock Deep learning for person re-identification: A survey and outlook.
\newblock \emph{IEEE Transactions on Pattern Analysis and Machine
  Intelligence}.

\bibitem[{Zhang et~al.(2019)Zhang, Lan, Zeng, and Chen}]{zhang2019densely}
Zhang, Z.; Lan, C.; Zeng, W.; and Chen, Z. 2019.
\newblock Densely semantically aligned person re-identification.
\newblock In \emph{Proceedings of the IEEE/CVF Conference on Computer Vision
  and Pattern Recognition}, 667--676.

\bibitem[{Zhang et~al.(2020)Zhang, Lan, Zeng, Jin, and
  Chen}]{zhang2020relation}
Zhang, Z.; Lan, C.; Zeng, W.; Jin, X.; and Chen, Z. 2020.
\newblock Relation-aware global attention for person re-identification.
\newblock In \emph{Proceedings of the ieee/cvf conference on computer vision
  and pattern recognition}, 3186--3195.

\bibitem[{Zhao et~al.(2021)Zhao, Jiang, Jia, Torr, and Koltun}]{zhao2021point}
Zhao, H.; Jiang, L.; Jia, J.; Torr, P.~H.; and Koltun, V. 2021.
\newblock Point transformer.
\newblock In \emph{Proceedings of the IEEE/CVF International Conference on
  Computer Vision}, 16259--16268.

\bibitem[{Zheng et~al.(2015)Zheng, Shen, Tian, Wang, Wang, and
  Tian}]{zheng2015scalable}
Zheng, L.; Shen, L.; Tian, L.; Wang, S.; Wang, J.; and Tian, Q. 2015.
\newblock Scalable person re-identification: A benchmark.
\newblock In \emph{Proceedings of the IEEE international conference on computer
  vision}, 1116--1124.

\bibitem[{Zheng et~al.(2017)Zheng, Zhang, Sun, Chandraker, Yang, and
  Tian}]{zheng2017person}
Zheng, L.; Zhang, H.; Sun, S.; Chandraker, M.; Yang, Y.; and Tian, Q. 2017.
\newblock Person re-identification in the wild.
\newblock In \emph{Proceedings of the IEEE Conference on Computer Vision and
  Pattern Recognition}, 1367--1376.

\bibitem[{Zheng et~al.(2019)Zheng, Karanam, Wu, and Radke}]{zheng2019re}
Zheng, M.; Karanam, S.; Wu, Z.; and Radke, R.~J. 2019.
\newblock Re-identification with consistent attentive siamese networks.
\newblock In \emph{Proceedings of the IEEE/CVF Conference on Computer Vision
  and Pattern Recognition}, 5735--5744.

\bibitem[{Zheng, Zheng, and Yang(2017)}]{zheng2017discriminatively}
Zheng, Z.; Zheng, L.; and Yang, Y. 2017.
\newblock A discriminatively learned cnn embedding for person reidentification.
\newblock \emph{ACM transactions on multimedia computing, communications, and
  applications (TOMM)}, 14(1): 1--20.

\bibitem[{Zhong et~al.(2020)Zhong, Zheng, Kang, Li, and Yang}]{zhong2020random}
Zhong, Z.; Zheng, L.; Kang, G.; Li, S.; and Yang, Y. 2020.
\newblock Random erasing data augmentation.
\newblock In \emph{Proceedings of the AAAI conference on artificial
  intelligence}, volume~34, 13001--13008.

\bibitem[{Zhu et~al.(2022)Zhu, Ke, Li, Liu, Tian, and Shan}]{zhu2022dual}
Zhu, H.; Ke, W.; Li, D.; Liu, J.; Tian, L.; and Shan, Y. 2022.
\newblock Dual Cross-Attention Learning for Fine-Grained Visual Categorization
  and Object Re-Identification.
\newblock In \emph{Proceedings of the IEEE/CVF Conference on Computer Vision
  and Pattern Recognition}, 4692--4702.

\bibitem[{Zhu et~al.(2020)Zhu, Guo, Liu, Tang, and Wang}]{zhu2020identity}
Zhu, K.; Guo, H.; Liu, Z.; Tang, M.; and Wang, J. 2020.
\newblock Identity-guided human semantic parsing for person re-identification.
\newblock In \emph{European Conference on Computer Vision}, 346--363. Springer.

\bibitem[{Zhu et~al.(2021)Zhu, Guo, Zhang, Wang, Huang, Qiao, Liu, Wang, and
  Tang}]{zhu2021aaformer}
Zhu, K.; Guo, H.; Zhang, S.; Wang, Y.; Huang, G.; Qiao, H.; Liu, J.; Wang, J.;
  and Tang, M. 2021.
\newblock AAformer: Auto-aligned transformer for person re-identification.
\newblock \emph{arXiv preprint arXiv:2104.00921}.

\end{thebibliography}

\end{document}